\def\PsyEvoArxiv{1}
\documentclass{article}
\usepackage{iclr2027_conference,times}
\usepackage{iftex}
\ifXeTeX
\usepackage{fontspec}
\IfFontExistsTF{Times New Roman}{\setmainfont{Times New Roman}}{\setmainfont{TeX Gyre Termes}}
\fi
\usepackage{amsmath,amssymb,amsthm}
\usepackage{graphicx}
\usepackage{placeins}
\usepackage{booktabs,tabularx}
\usepackage{algorithm,algpseudocode}
\usepackage{url}
\usepackage[hidelinks]{hyperref}

\newtheorem{proposition}{Proposition}

\title{PsyEvo: A Personalized Counseling Agent That Self-Evolves at Test Time}
\ifdefined\PsyEvoArxiv
\usepackage{xcolor}
\iclrfinalcopy
\author{%
\normalfont Yuting Yan\textsuperscript{1,*}, Shihao Xu\textsuperscript{1,2,*}, Junhao Yu\textsuperscript{1}, Mingcong Zuo\textsuperscript{1}\\
Lu Chen\textsuperscript{1}, Nan Xiang\textsuperscript{1}, Haiyang Geng\textsuperscript{1,2}, Dongjie Tao\textsuperscript{1}, Minghao Wang\textsuperscript{1}\\[0.5em]
\normalfont\textsuperscript{1}Lyncia Lab\\
\normalfont\textsuperscript{2}Tianqiao and Chrissy Chen Institute\\[0.3em]
\normalfont\small\textsuperscript{*}Equal contribution.
}
\hypersetup{pdftitle={PsyEvo: A Personalized Counseling Agent That Self-Evolves at Test Time},
pdfauthor={Yuting Yan, Shihao Xu, Junhao Yu, Mingcong Zuo, Lu Chen, Nan Xiang, Haiyang Geng, Dongjie Tao, Minghao Wang}}
\else
\author{Anonymous}
\fi

\begin{document}
\maketitle
\ifdefined\PsyEvoArxiv
\lhead{\sffamily\fontsize{9}{11}\selectfont\color[gray]{0.38}PSYEVO\enspace\textperiodcentered\enspace PREPRINT}
\renewcommand{\headrulewidth}{0.25pt}
\renewcommand{\headrule}{\hbox to\headwidth{\color[gray]{0.72}\leaders\hrule height\headrulewidth\hfill}\vskip-\headrulewidth}
\fi
\begin{abstract}
Mental health disorders affect a substantial proportion of the global population, yet a persistent shortage of trained practitioners leaves the majority without adequate care.
Large language model (LLM)-based counselors present a promising direction for delivering scalable conversational psychological support.
Offline model training alone leaves limited room to adapt to individual clients or to learn from ongoing therapeutic interaction at test time.
We introduce PsyEvo, an LLM-based counseling framework that enables both client-specific personalization and response-policy improvement at test time through three components:
Hierarchical Bayesian Skill Policy (HBSP) personalizes what intervention to apply by maintaining a per-client skill posterior updated from session feedback;
Inter-session Listwise Preference Optimization (LiPO) improves how the selected skill is expressed by updating a shared response adapter from cross-client preference evidence;
and State-conditioned Ordinal Credit Assignment (SOCA) supplies candidate preferences and trajectory credit to the two components through consistency-checked comparisons and ordinal projection.
In simulated-client evaluation with shared online cohort adaptation, PsyEvo obtains 7.684 Overall on PsychEval and exceeds every component variant in each of three matched runs.
Removing individual components lowers mean overall score by 0.138--0.171 under the shared configuration, supporting conditional contributions within the complete scaffold.
\ifdefined\PsyEvoArxiv
Our code is available at \url{https://github.com/Lingxi-mental-health/PsyEvo}.
\fi
\end{abstract}

\noindent Keywords: test-time learning; multi-session counseling; personalized counseling; ordinal credit assignment; preference optimization; self-evolving agents.

\section{Introduction}
Mental health disorders have become a major global health challenge, with nearly one billion individuals affected and a substantial share of the worldwide disease burden attributed to conditions such as depression and anxiety \citep{whoDepression,whoAnxiety}.
Psychological counseling is a widely adopted evidence-based intervention, yet training qualified counselors requires years of supervision, and the available workforce falls far short of demand \citep{bacpTraining,who2022mental}.
In this context, large language models (LLMs) provide a scalable conversational interface that could expand access to psychological support.

Research on LLM-based counseling spans supervised fine-tuning (SFT), preference optimization, and agent-based mechanisms.
SoulChat and CPsyCoun provide counseling and empathetic dialogue data for SFT, whereas PsyCoPref supplies preference comparisons for response optimization \citep{chen2023soulchat,zhang2024cpsycoun,zhang2026psycopref}.
MusPsy and PsychAgent incorporate memory, planning, and skill-related mechanisms to support longitudinal interaction \citep{wang2026muspsy,yang2026psychagent}.
These approaches extend counseling support from individual responses to multiple sessions.
Beyond continuity, individual clients can differ in their experiences, needs, and communication styles \citep{wang2026muspsy}, motivating updates to intervention selection and to the response policy during test-time interaction.
This motivates two complementary learning objectives: personalization, which tailors intervention selection to a particular client, and counselor self-evolution, which updates the response policy using cross-client experience.

To this end, we present PsyEvo, a test-time learning framework for multi-session psychological counseling that addresses both objectives over a frozen backbone through three components: Hierarchical Bayesian Skill Policy (HBSP) for personalized intervention selection via a per-client skill posterior, Listwise Preference Optimization (LiPO) for shared response-policy improvement via cross-client preference graphs, and State-conditioned Ordinal Credit Assignment (SOCA) for constructing candidate preferences and trajectory credit from relative comparisons and realized feedback.
Our contributions are as follows: (1) we introduce PsyEvo, which combines client-private skill posteriors with a shared population response adapter over a frozen backbone, using HBSP for personalized skill selection and LiPO for cross-session response-policy learning; (2) we propose SOCA, which converts compressed self-evaluation feedback into candidate preferences through consistency-checked comparisons and into delayed credit through within-trajectory ordinal propagation; and (3) we establish positive conditional component contributions across three matched PsychEval runs and evaluate a related evidence-fusion rule on SummEval, where it achieves the highest mean agreement with human judgments among the compared evaluators.

\begin{figure*}[t]
\centering
\includegraphics[width=\textwidth]{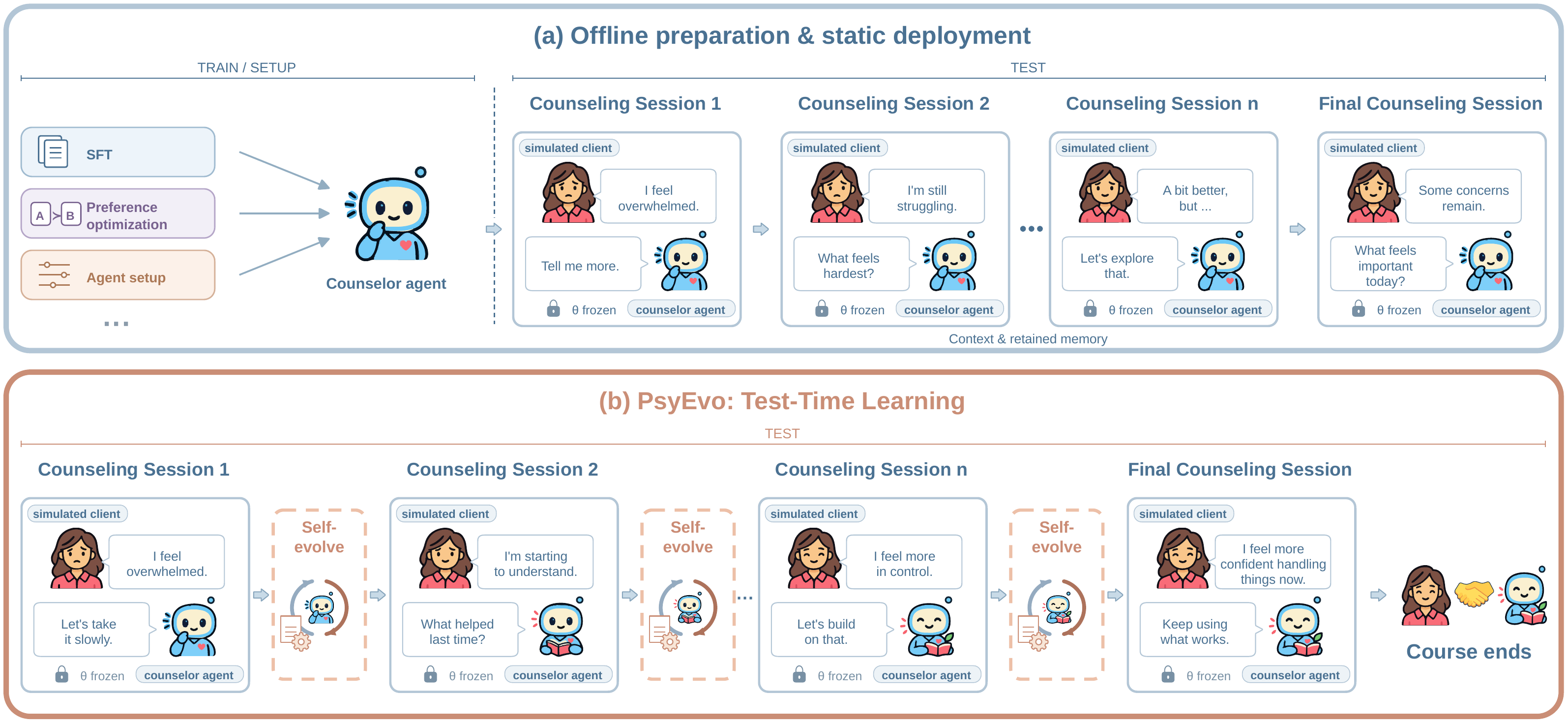}
\caption{Static deployment versus test-time learning in psychological counseling.
(a) A statically deployed counselor keeps its parameters fixed across sessions; (b) PsyEvo keeps the backbone frozen, updates the client-specific skill posterior after each session, and schedules shared-adapter updates at five-session intervals on PsychEval, subject to sufficient preference evidence.}
\label{fig:paradigm}
\end{figure*}

\section{PsyEvo}
\label{sec:method}

%This section first formalizes multi-session counseling as a sequential decision process and decomposes the counselor's state into client-private and population-shared components.
%We then present the three components of PsyEvo: HBSP for personalized skill selection (Section~\ref{sec:hbsp}), LiPO for cross-client response-policy improvement (Section~\ref{sec:lipo}), and SOCA for ordinal credit construction (Section~\ref{sec:soca}).

\subsection{Problem Formulation}

We formalize multi-session psychological counseling as a sequential decision process.
Let $u \in \mathcal{U}$ denote a client, $k$ the session index, and $t$ the turn index within a session.
At each turn, the counselor observes a state $s_t$ comprising the realized dialogue history, the inferred client state, the therapeutic modality, the course stage, and a cross-session memory $M_u$.
A skill controller $\mu_u$ selects a counseling skill $a_t$ from the set $\mathcal{A}_t$ of skills valid for the current modality, and the response policy $\pi_{\theta,\phi^{(v)}}$ generates $N$ candidate responses $y_t^{(1)},\ldots,y_t^{(N)}$ conditioned on $s_t$ and $a_t$; one candidate is delivered to the client.

PsyEvo decomposes the counselor's learning state into a client-private component and a population-shared component:
\begin{equation}
\Omega_u=(M_u,b_u),\qquad
\Psi=(\phi^{(v)},\mathcal B_{\mathrm{pop}}),\qquad
\pi_u(a,y\mid s)=\mu_u(a\mid s)\pi_{\theta,\phi^{(v)}}(y\mid s,a),
\label{eq:state}
\end{equation}
where the backbone parameters $\theta$ remain frozen throughout.
The private state $\Omega_u$ consists of the client's memory $M_u$ and a skill posterior $b_u$; the population state $\Psi$ contains the versioned shared adapter $\phi^{(v)}$ and a population evidence buffer $\mathcal{B}_{\mathrm{pop}}$.

\begin{figure*}[t]
\centering
\includegraphics[width=\textwidth]{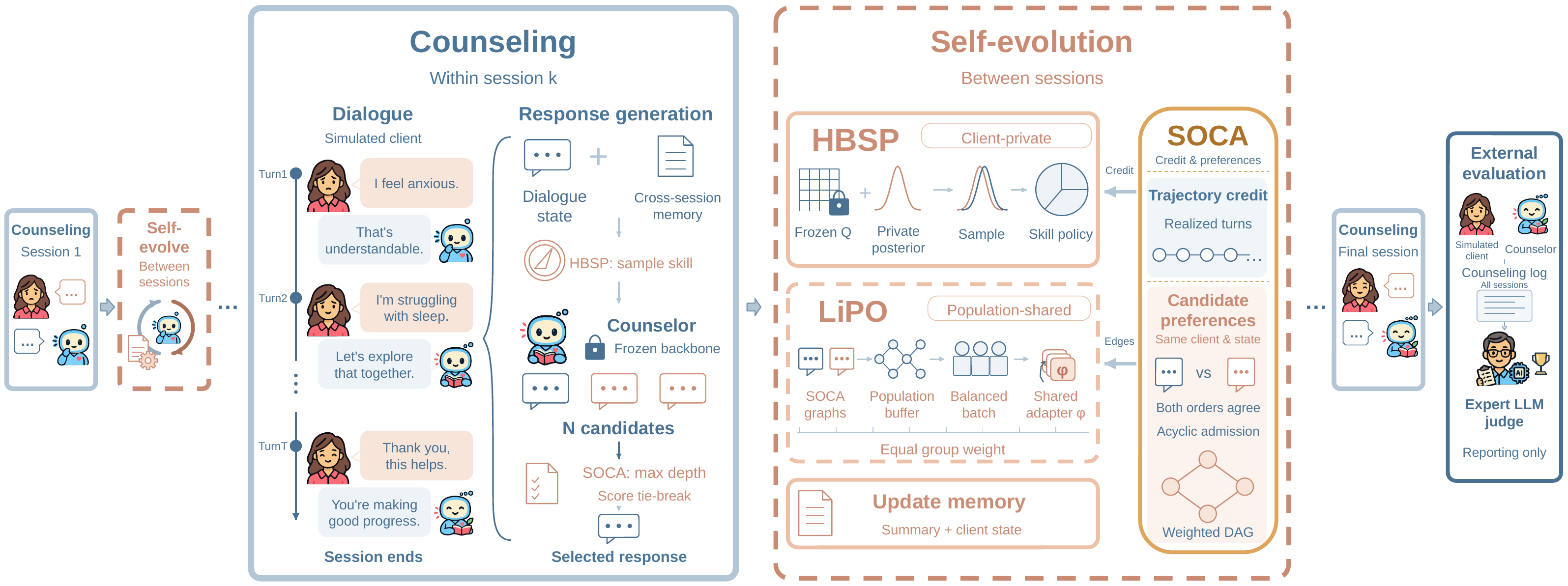}
\caption{Overview of PsyEvo's test-time learning cycle.
Within a session, HBSP selects a skill and SOCA selects the delivered response; between sessions, HBSP updates the client-private posterior using SOCA trajectory credit, while LiPO periodically updates the shared adapter from SOCA preference graphs accumulated across clients.
External judges score completed transcripts for reporting only and do not feed the learning signals.}
\label{fig:pipeline}
\end{figure*}

Three components operate over this decomposition (Figure~\ref{fig:pipeline}).
HBSP updates $b_u$ to personalize which skill to select for each client; LiPO updates $\phi^{(v)}$ to improve how the selected skill is expressed; and SOCA constructs the learning signals, i.e. candidate preferences and trajectory credit, consumed by both.

\subsection{HBSP}
\label{sec:hbsp}

First, we propose HBSP to determine which counseling skill to apply at each turn.
A shared skill-prior table $Q(s,a)$, initialized from skill-labeled PsychEval dialogue, provides a common starting point for all clients.
During interaction, $Q$ remains fixed; only the client-private posterior $b_u(s,a)$ is updated to capture individual preferences.
Each client starts with $m_u(s,a)=0$ and $\lambda_u(s,a)=4$, corresponding to a residual prior variance of $0.25$; we use observation variance $\sigma_b^2=1$.

Concretely, HBSP decomposes the skill-selection score into the shared term and a client-specific residual drawn from a normal posterior:
\begin{align}
Q_u(s,a)&=Q(s,a)+b_u(s,a),\qquad b_u(s,a)\sim\mathcal N(m_u(s,a),\lambda_u(s,a)^{-1}),\\
p_t(a)&=(1-\varepsilon_t)\mathrm{softmax}\!\left(\tau[Q(s_t,a)+\tilde b_u(s_t,a)]\right)
+\varepsilon_t/|\mathcal A_t|.
\label{eq:hbsp-policy}
\end{align}
Here, $\tilde b_u(s_t,a)$ denotes a draw from $\mathcal N(m_u(s_t,a),\lambda_u(s_t,a)^{-1})$ for the current skill decision.
At the end of each session, SOCA assigns a trajectory return $G_t$ to the skill selected at turn $t$ (Section~\ref{sec:soca}).
With residual $d_t=G_t-Q(s_t,a_t)$, HBSP uses the Gaussian working likelihood $d_t\mid b_u(s_t,a_t)\sim\mathcal N(b_u(s_t,a_t),\sigma_b^2)$ to update the posterior:
\begin{equation}
\lambda'_u=\lambda_u+\sigma_b^{-2},\qquad
m'_u=\frac{\lambda_um_u+\sigma_b^{-2}d_t}{\lambda'_u}.
\label{eq:hbsp-update}
\end{equation}
Posterior sampling uses model-based uncertainty for exploration: repeated observations increase precision, while less-visited state--skill cells retain wider sampling distributions. Appendix~\ref{app:hbsp-shrinkage} derives how the shared starting score and observed returns combine under this update.

\subsection{LiPO}
\label{sec:lipo}

While HBSP personalizes skill selection per client, we employ LiPO to improve how the selected skill is expressed by updating a shared response adapter from preference evidence accumulated across clients.
For each candidate group $g$ generated for client $u$, SOCA provides a set of supported preference edges $E_{u,g}$ with positive weights $w_{ij}$ (Section~\ref{sec:soca}).
Each graph retains its client and modality identifiers; no edge is constructed between candidates from different clients.

We utilize a shared rank-8 LoRA adapter \citep{hu2022lora} as the learnable module.
For a preference edge $i \succ j$, the policy--reference log-ratio utility $h_{u,g,i}$ and the per-group loss are
\begin{equation}
h_{u,g,i}=\beta\log\frac{\pi_{\theta,\phi^{(v)}}(y_i\mid s,a)}{\pi_{\theta,0}(y_i\mid s,a)},\qquad
\mathcal L_{u,g}=\frac1{|E_{u,g}|}\sum_{(i,j)\in E_{u,g}}w_{ij}
\operatorname{softplus}[-(h_{u,g,i}-h_{u,g,j})].
\label{eq:lipo}
\end{equation}
Our objective averages SOCA-weighted, DPO-style logistic losses within each candidate group.
For a single edge with unit weight, Eq.~\ref{eq:lipo} reduces to the standard pairwise DPO loss. LiPO retains the candidate-group structure, weights each supported edge by its bidirectional SOCA support strength, and normalizes by the number of retained edges in that group; pairwise DPO optimizes individual preferred--dispreferred pairs with unit weight.
The population objective averages the per-group losses over the candidate groups in the training batch, so that a group does not receive greater weight merely because it contains more preference edges:
\begin{equation}
\mathcal L_{\mathrm{pop}}=
\frac{1}{\sum_{u\in\mathcal U_B}|\mathcal G_u|}
\sum_{u\in\mathcal U_B}\sum_{g\in\mathcal G_u}\mathcal L_{u,g}.
\label{eq:lipo-pop}
\end{equation}
The reference $\pi_{\theta,0}$ is the frozen backbone with the adapter disabled.
An accepted adapter version becomes available only to sessions that start after the update boundary, ensuring that response generation uses a fixed version throughout each session.

\subsection{SOCA}
\label{sec:soca}

Both HBSP and LiPO require informative learning signals, yet model self-evaluation can assign nearly identical scores to plausible counseling responses, a problem we call score compression.
To address this issue, we introduce SOCA to construct evidence at two resolutions: candidate preferences for LiPO from pairwise comparisons within a state, and trajectory credit for HBSP from ordinal projection along the realized session.

\paragraph{Candidate preferences.}
For each unordered candidate pair $\{i,j\}$, let $p_{ij}^{\rightarrow}$ and $p_{ij}^{\leftarrow}$ denote weighted judgment support for candidate $i$ under the two presentation orders.
SOCA accepts a direction only when both orders agree; otherwise it abstains.
The direction-consistent edge strength is
\begin{equation}
c_{ij}=\begin{cases}
2\min(p_{ij}^{\rightarrow},p_{ij}^{\leftarrow})-1, & i\succ j,\\
2\min(1-p_{ij}^{\rightarrow},1-p_{ij}^{\leftarrow})-1, & j\succ i.
\end{cases}
\label{eq:soca-graph}
\end{equation}
Edges below a stability floor are discarded.
The remaining edges are added in decreasing strength, skipping any that would create a directed cycle; their strengths serve as the LiPO weights $w_{ij}$ in Eq.~\ref{eq:lipo}.
We orient each edge from the preferred candidate to the dispreferred candidate and define depth as the longest directed path from a candidate to a sink, with sink depth zero.
The delivered response maximizes this depth; ties are resolved by the higher pointwise score and then the smaller candidate index.

\paragraph{Trajectory credit.}
For the realized trajectory, let $R_t$ be the turn outcome and $\rho_t$ its tie-aware midrank among the $T$ session turns.
SOCA maps midranks to ordinal levels, scales them by $s=0.5$, and applies a mean-preserving shift $\xi$:
\begin{equation}
\tilde r_t=\mathrm{clip}(\bar R+s\ell_t+\xi,0,1),
\qquad \frac1T\sum_{t=1}^{T}\tilde r_t=\bar R=\frac1T\sum_{t=1}^{T}R_t.
\label{eq:ordinal}
\end{equation}
For zero-based midranks, we use $K=\min(3,T)$ and $\ell_t=2\min(K-1,\lfloor K\rho_t/T\rfloor)/(K-1)-1$ when $T>1$, and $\ell_1=0$ otherwise.
Equal scores receive equal midranks and therefore equal projected credit.
Discounted propagation then connects earlier skill decisions with later outcomes:
\begin{equation}
G_t=\tilde r_t+\gamma G_{t+1},\qquad G_{T+1}=r_{\mathrm{session}},\qquad\gamma=0.9.
\label{eq:return}
\end{equation}
HBSP receives returns from the realized trajectory, while LiPO receives preferences from candidate comparisons.

Algorithm~\ref{alg:lifecycle} summarizes the interaction and update cycle.
Posterior and adapter updates take effect at session boundaries, keeping their values fixed within each session.

\begin{algorithm}[htbp]
\caption{Test-time learning cycle of PsyEvo. All parameter updates take effect only at session boundaries.}
\label{alg:lifecycle}
\begin{algorithmic}[1]
\Require Frozen backbone $\theta$, shared skill table $Q$, adapter $\phi^{(v)}$, population buffer $\mathcal B_{\mathrm{pop}}$
\Require Client $u$ with private memory $M_u$ and skill posterior $b_u$
\For{each session $(u,k)$}
\State Fix adapter version $\phi^{(v)}$ for the current session
\Comment{Within-session loop}
\For{each turn $t = 1, \ldots, T$}
\State Construct dialogue state $s_t$ from history and $M_u$
\State Sample skill $a_t \sim p_t(\cdot)$ via HBSP posterior (Eq.~\ref{eq:hbsp-policy})
\State Generate $N$ candidates $\{y_t^{(1)}, \ldots, y_t^{(N)}\}$ from $\pi_{\theta,\phi^{(v)}}(\cdot \mid s_t, a_t)$
\State Evaluate candidates via pointwise scoring and two-order pairwise comparisons
\State Construct acyclic SOCA preference graph (Eq.~\ref{eq:soca-graph}); select response by graph depth
\State Deliver selected response; record client reaction
\State Store local graph with client, modality, and version metadata in $\mathcal B_{\mathrm{pop}}$
\EndFor
\Comment{Between-session updates}
\State Compute ordinal trajectory credit $\{G_t\}_{t=1}^{T}$ via SOCA (Eqs.~\ref{eq:ordinal}--\ref{eq:return})
\State Update client posterior $b_u$ using realized residuals (Eq.~\ref{eq:hbsp-update})
\If{population update criterion is met}
\State Sample balanced batch from $\mathcal B_{\mathrm{pop}}$ and normalize losses within each candidate group
\State Update shared adapter: $\phi^{(v)} \rightarrow \phi^{(v+1)}$ by optimizing $\mathcal L_{\mathrm{pop}}$ (Eq.~\ref{eq:lipo-pop})
\State Rollback to $\phi^{(v)}$ on optimization failure
\EndIf
\State Update cross-session memory $M_u$ from the realized transcript
\EndFor
\end{algorithmic}
\end{algorithm}

\section{Experiments}

%We evaluate PsyEvo on multi-session counseling quality and assess SOCA's evidence-aggregation quality against human judgments.
%In this section, we will introduce the datasets and evaluation metrics, the baselines and implementation details, the main experimental results, and the longitudinal and ablation analyses, respectively.

\subsection{Datasets and Baselines}
\label{sec:datasets}
\label{sec:baselines-implementation}
Our main counseling study uses a 100-client PsychEval cohort across five therapy modalities \citep{pan2026psycheval}; SummEval \citep{fabbri2021summeval} tests evidence aggregation beyond counseling against human judgments. Supplementary MusPsy figures for 100 profiles over six sessions are retained in Appendix~\ref{app:muspsy-results} \citep{wang2026muspsy}.
For counseling, we compare PsyEvo with eleven baselines, seven counseling-specific systems and four general-purpose models (Table~\ref{tab:counseling-main}), and with a Qwen3-32B variant trained with DPO (Table~\ref{tab:components}); for summary evaluation, we compare SOCA with six evidence-aggregation baselines that share its Qwen3-32B backbone.
PsyEvo uses a Qwen3-32B counseling backbone with a rank-8 LoRA adapter optimized by LiPO. Internal candidate judgments also use Qwen3-32B; external PsychEval reporting uses Gemini-3-Flash on completed transcripts (Appendix~\ref{app:judgesfull}).
PsychEval follows the PsychAgent interaction protocol, with BGE-M3 retrieval and DeepSeek-V3.1 client simulation; MusPsy follows its benchmark-specific client-state, six-session interaction, and evaluation protocols \citep{yang2026psychagent,wang2026muspsy}.
In PsychEval, PsyEvo updates its shared adapter across clients in the online evaluation cohort, while each private skill posterior uses only that client's feedback.
Appendices~\ref{app:datasets-detail},~\ref{app:baselines-detail},~\ref{app:implementationfull}, and~\ref{app:costfull} detail the datasets, baselines, learning settings, and training costs.

\subsection{Evaluation Metrics}
\label{sec:metrics}
On PsychEval, we report four metric families derived from its 18 evaluation instruments \citep{pan2026psycheval}.
Counselor-shared (C-Shared) summarizes common dimensions of counseling quality, including working alliance and relational authenticity.
Counselor-specific (C-Spec) uses CTRS, MITI, TES, PSC, and EFT-TFS for CBT, behavioral, humanistic, psychodynamic, and postmodernist therapy, respectively.
Client-shared (L-Shared) summarizes common simulated-client scores, including affect and session alliance; client-specific (L-Spec) uses BDI-II, STAI, CCT, IPO, and SFBT indicators for the corresponding modalities.
Metric definitions are detailed in Appendix~\ref{app:metrics-detail}.
Overall is the unweighted average of the four reported family scores.

For MusPsy, we report the Working Alliance Inventory--Short Form (WAI; Goal, Task, Bond, and their average) and positive and negative PANAS affect scores on a 1--5 scale \citep{wang2026muspsy}, averaged over six sessions and 100 clients using the official MusPsy evaluation code and judging prompt.
While for SummEval, we compute Spearman's $\rho$ between automatic and human ratings for coherence (COH), consistency (CON), fluency (FLU), and relevance (REL) within each source-document group and macro-average across groups \citep{fabbri2021summeval}; Mean averages the four dimensions.
Higher values are better for all metrics except negative PANAS.
Appendix~\ref{app:metrics-detail} lists the constituent instruments, aggregation rules, and judging-prompt protocols.

\subsection{Main Results}
\label{sec:main-results}
%We evaluate both end-to-end counseling quality, where PsyEvo surpasses all listed baselines on PsychEval and MusPsy, and evidence aggregation on SummEval, where SOCA achieves higher agreement with human judgments than the compared scoring methods.

\paragraph{Multi-Session Counseling Quality}
\label{sec:psyevo-results}

\begin{table}[ht]
\centering\small
\caption{PsychEval results. P denotes a published reference score; E denotes our evaluation. PsyEvo averages seeds 7, 17, and 27 under shared online cohort adaptation. All metrics favor higher values. Complete-system rankings are descriptive; the matched component contrasts appear in Table~\ref{tab:components}.}
\label{tab:counseling-main}
\label{tab:pa100}
\setlength{\tabcolsep}{4pt}
\begin{tabular}{@{}llrrrrr@{}}
\toprule
Method & Source & Overall & C-Shared & C-Spec & L-Shared & L-Spec\\
\midrule
CACTUS & E & 4.869 & 5.112 & 2.530 & 4.594 & 7.240\\
CPsyCounX & P & 4.680 & 4.210 & 2.510 & 4.730 & 7.270\\
SMILE & E & 4.142 & 4.261 & 2.658 & 3.794 & 5.856\\
PsyDTLLM & P & 6.055 & 6.100 & 5.430 & 5.270 & 7.420\\
PsyLLM-thinking-v3 & P & 5.883 & 5.300 & 4.670 & 5.630 & 7.930\\
TheraMind & P & 6.625 & 6.250 & 6.940 & 5.480 & 7.830\\
PsychAgent-32B & P & 7.348 & 7.320 & 7.910 & 5.920 & 8.240\\
GPT-5.6-terra & E & 6.839 & 7.297 & 7.142 & 5.906 & 7.012\\
DeepSeek-V4-Flash & E & 6.555 & 6.543 & 5.917 & 5.891 & 7.872\\
Gemini-3.8-Flash & E & 7.363 & 7.310 & 8.150 & 6.224 & 7.770\\
Qwen3-32B & E & 7.028 & 6.851 & 7.012 & 6.214 & 8.033\\
\textbf{PsyEvo} & E & \textbf{7.684} & \textbf{7.527} & \textbf{8.260} & \textbf{6.468} & \textbf{8.481}\\
\bottomrule
\end{tabular}
\end{table}

As shown in Table~\ref{tab:counseling-main}, PsyEvo achieves an Overall score of 7.684 on PsychEval. The descriptive gap is 0.336 over the strongest published counseling-specific reference, PsychAgent-32B (7.348), and 0.321 over the strongest general-purpose model evaluated by us, Gemini-3.8-Flash (7.363).
Among the four metric families, L-Shared shows the largest improvement over PsychAgent-32B (+0.548), followed by C-Spec (+0.350); C-Shared and L-Spec improve by 0.207 and 0.241, respectively.

The supplementary MusPsy comparison is recorded separately in Appendix~\ref{app:muspsy-results}, where its unresolved main-run configuration linkage is stated.

\paragraph{Summary Evaluation with SOCA}
\label{sec:soca-results}
We evaluate a SummEval variant of SOCA that combines pointwise, pairwise, and repair evidence to rank summaries against human quality judgments.
This fusion variant shares pointwise and bidirectional comparison evidence with counseling SOCA, but does not test its graph-based response selector or ordinal trajectory credit, nor HBSP or LiPO; the component-removal analysis in Section~\ref{sec:soca-ablation} examines the contribution of each evidence channel.
As shown in Table~\ref{tab:summ}, SOCA achieves the highest Spearman correlation with human judgments on coherence, fluency, and relevance, and ties for the highest consistency correlation on the full SummEval dataset.
Its mean correlation of 0.508 surpasses the next-best method PCA (0.489) by 0.019 and JDM (0.467) by 0.041, with the largest gains on fluency (+0.036 over PCA) and relevance (+0.015 over PCA).

\begin{table}[htbp]
\centering
\footnotesize
\caption{Correlation with human judgments on SummEval.
Values are within-document Spearman correlations between automatic and human scores for coherence (COH), consistency (CON), fluency (FLU), and relevance (REL), computed from all 1,600 summaries grouped into 100 source documents and macro-averaged across documents, with all methods using a Qwen3-32B backbone.}
\label{tab:summ}
\label{tab:summdev}
\begin{tabular}{lrrrrr}
\toprule
Method & COH $\uparrow$ & CON $\uparrow$ & FLU $\uparrow$ & REL $\uparrow$ & Mean $\uparrow$\\
\midrule
Raw Scalar & 0.440 & 0.449 & 0.374 & 0.443 & 0.427\\
JDM & 0.544 & 0.431 & 0.412 & 0.479 & 0.467\\
PCA & 0.590 & 0.436 & 0.412 & 0.520 & 0.489\\
PairS-Greedy & 0.558 & 0.390 & 0.366 & 0.419 & 0.433\\
PairS-Beam & 0.553 & 0.388 & 0.356 & 0.417 & 0.429\\
BatchEval & 0.497 & 0.422 & 0.358 & 0.500 & 0.444\\
\midrule
\textbf{SOCA} & \textbf{0.600} & \textbf{0.449} & \textbf{0.448} & \textbf{0.535} & \textbf{0.508}\\
\bottomrule
\end{tabular}
\end{table}

\subsection{Ablation Study}
\label{sec:ablations}

\paragraph{PsyEvo Component Ablations}
We ablate each component conditional on the remaining scaffold (Table~\ref{tab:components}); these contrasts do not decompose the full advantage over the base model.
The SOCA comparison jointly tests candidate-preference construction and trajectory-credit projection. Table~\ref{tab:ablation-contract} specifies the changes and retained learning paths for each comparison.
PsyEvo exceeds every component variant in Overall in each of the three matched runs. Disabling LiPO-based parameter updates reduces the mean Overall score by 0.171; removing SOCA reduces it by 0.152, replacing LiPO with pairwise DPO by 0.146, and removing HBSP by 0.138.
The largest decreases for removing LiPO and SOCA occur in L-Spec (0.429 and 0.383, respectively), while removing HBSP most affects L-Spec (0.367).
Appendix~\ref{app:seed-results} reports individual seeds, standard deviations, and paired differences whose conditional client-bootstrap 95\% intervals are above zero for all four comparisons.

\begin{table}[ht]
\centering\small
\caption{Ablation of PsyEvo's three components on PsychEval.
All variants share the experimental configuration, with only the specified component removed or replaced; with DPO replaces the weighted candidate-group objective with pairwise DPO, w/o LiPO disables LiPO-based parameter updates, and Overall averages the four metric families. Scores are means across seeds 7, 17, and 27.}
\label{tab:components}
\begin{tabular}{lrrrrr}
\toprule
Variant & C-Shared & C-Spec & L-Shared & L-Spec & Overall\\
\midrule
PsyEvo w/o SOCA & 7.510 & 8.010 & 6.510 & 8.098 & 7.532\\
PsyEvo with DPO & 7.350 & 8.300 & 6.450 & 8.052 & 7.538\\
PsyEvo w/o LiPO & 7.470 & 7.990 & \textbf{6.540} & 8.052 & 7.513\\
PsyEvo w/o HBSP & 7.320 & \textbf{8.400} & 6.350 & 8.114 & 7.546\\
\textbf{PsyEvo} & \textbf{7.527} & 8.260 & 6.468 & \textbf{8.481} & \textbf{7.684}\\
\bottomrule
\end{tabular}
\end{table}

\paragraph{SOCA Component Ablations}
\label{sec:soca-ablation}
We separately remove each of SOCA's three evidence channels---pointwise scoring, pairwise comparisons, and repair evidence---on the full SummEval dataset (see Table~\ref{tab:soca-ablation}).
Removing pairwise comparisons produces the largest drop in mean correlation ($-$0.041), with coherence most affected ($-$0.089).
We also found that removing pointwise scoring or repair evidence leads to smaller decreases of 0.016 and 0.013, respectively.

\begin{table}[htbp]
\centering\small
\caption{Ablation of SOCA's evidence channels on SummEval.
Each variant removes one evidence channel; values are within-document Spearman correlations with human judgments, macro-averaged across the full SummEval dataset under a Qwen3-32B backbone.}
\label{tab:soca-ablation}
\setlength{\tabcolsep}{3.5pt}
\begin{tabular}{@{}lrrrrr@{}}
\toprule
Method & COH $\uparrow$ & CON $\uparrow$ & FLU $\uparrow$ & REL $\uparrow$ & Mean $\uparrow$\\
\midrule
SOCA w/o Pointwise Scoring & 0.581 & 0.434 & 0.432 & 0.522 & 0.492\\
SOCA w/o Pairwise Comparisons & 0.511 & 0.439 & 0.438 & 0.483 & 0.467\\
SOCA w/o Repair Evidence & 0.587 & 0.439 & 0.422 & 0.529 & 0.495\\
\textbf{SOCA} & \textbf{0.600} & \textbf{0.449} & \textbf{0.448} & \textbf{0.535} & \textbf{0.508}\\
\bottomrule
\end{tabular}
\end{table}

\subsection{Human Evaluation}
\label{sec:human-evaluation}
In a descriptive pilot, three psychology experts blind-rated simulated dialogues from 25 cases (three sessions each) on five session-level dimensions and case-level client improvement, using a 1--5 scale (Appendix~\ref{app:human-evaluation}).
PsyEvo achieves the highest mean overall rating of 4.10, exceeding PsychAgent-32B (3.58) by 0.52 and CPsyCounX (2.95) by 1.15 (Table~\ref{tab:human-main}).
These are descriptive means from a small pilot with low dimension-level agreement (Appendix~\ref{app:human-agreement}); client improvement denotes impressions of simulated dialogue, not clinical change or established expert consensus.
Composite absolute-agreement reliability is ICC(A,3)=0.580 (95\% CI [0.398, 0.690]).

\begin{table}[htbp]
\centering\scriptsize
\caption{Descriptive pilot ratings of three counseling systems on 25 cases.
Three psychology experts blind-rated three sessions per case on five session-level dimensions and case-level client improvement (1--5); Overall is the unweighted mean of the six dimension means.}
\label{tab:human-main}
\setlength{\tabcolsep}{3pt}
\resizebox{\linewidth}{!}{%
\begin{tabular}{lrrrrrrr}\toprule
System & Ethics & \shortstack{Empathy\\accuracy} & \shortstack{Empathy\\quality} & Interaction & Intervention & \shortstack{Client\\improvement} & Overall\\\midrule
PsyEvo & \textbf{4.53} & \textbf{3.89} & \textbf{3.73} & \textbf{4.11} & \textbf{4.13} & \textbf{4.20} & \textbf{4.10}\\
PsychAgent-32B & 4.49 & 3.49 & 3.22 & 3.47 & 3.67 & 3.13 & 3.58\\
CPsyCounX & 4.49 & 2.76 & 2.64 & 2.49 & 2.84 & 2.47 & 2.95\\\bottomrule
\end{tabular}}
\end{table}

\subsection{Analysis}
\label{sec:interpretability}
We describe client-specific skill rankings, session-wise outcomes, and a longitudinal case to examine learning-state changes alongside counseling behavior.

\begin{figure}[htbp]
\centering\includegraphics[width=0.8\linewidth]{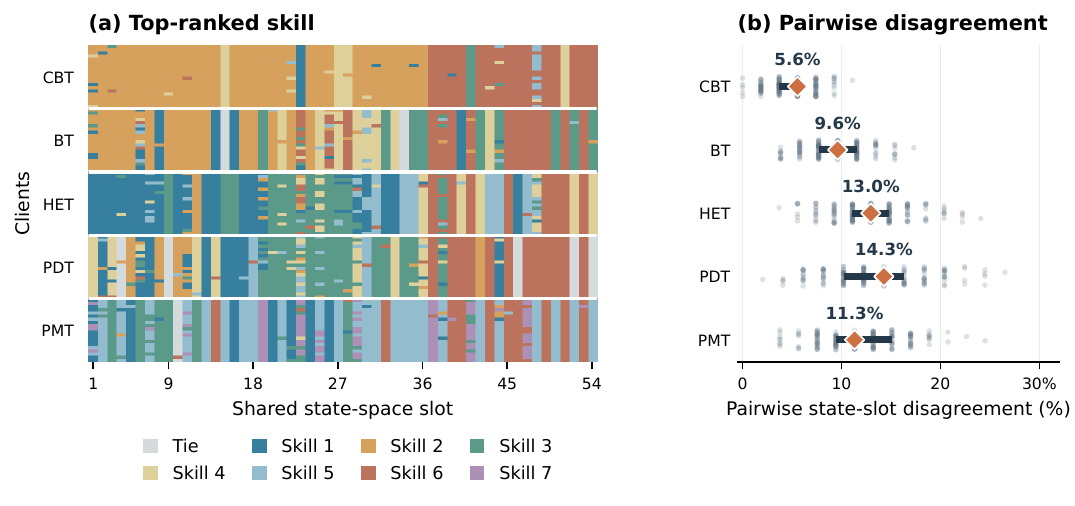}
\caption{Client-specific HBSP skill rankings at shared states on PsychEval.
(a) Top-ranked skill under HBSP for 20 clients per modality across 54 state-space slots; gray cells mark ties.
(b) Pairwise disagreement fraction over the 190 client pairs per modality; diamonds mark medians and horizontal lines the interquartile range.}
\label{fig:hbsp-personalization}
\end{figure}

\paragraph{Client-Specific Skill Personalization}
\label{sec:personalization}
We analyze HBSP skill rankings on PsychEval for 100 clients across five modalities to examine how skill preferences differ between clients at shared states.
At the same state-space slot, different clients can receive different top-ranked skills (Figure~\ref{fig:hbsp-personalization}): within-modality median pairwise disagreement ranges from 5.6\% (CBT) to 14.3\% (PDT), while most slots agree.
These differences document client-specific ranking adaptation; the ablation (Table~\ref{tab:components}) tests its conditional contribution, but does not isolate client matching from residual learning (Appendix~\ref{app:mechanismsfull}).
Separate archived adapter diagnostics are given in Appendix~\ref{app:mechanismsfull}; their unverified connection to the main runs limits them to illustrating that trace.

\begin{figure}[htbp]
\centering\includegraphics[width=0.8\linewidth]{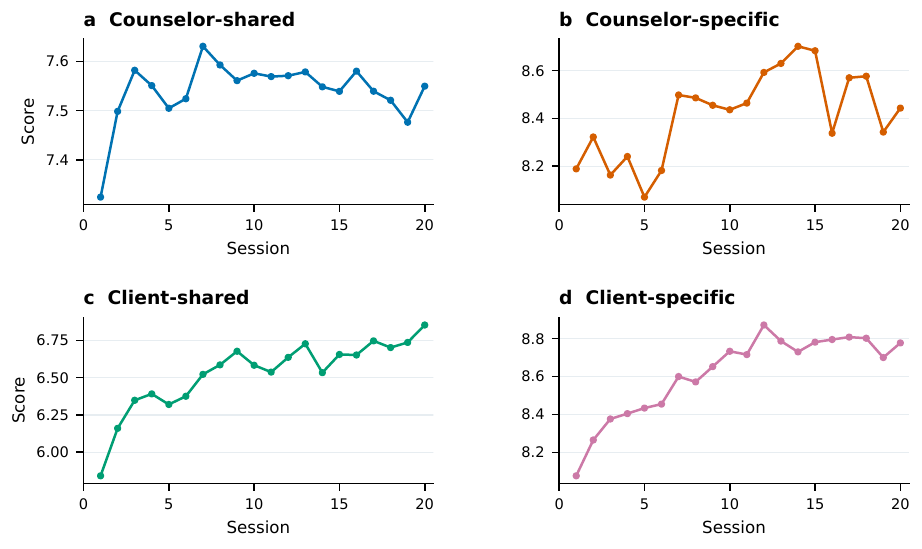}
\caption{PsychEval score trajectories over 20 sessions.
(a--d) Curves show session-wise mean counselor-shared, counselor-specific, client-shared, and client-specific scores, respectively.
The number of scored clients varies across sessions, and panels use separate vertical ranges.
Higher scores are better for all metrics.}
\label{fig:session-trajectories}
\end{figure}

\paragraph{Session-wise Outcomes}
\label{sec:longitudinal}
Figure~\ref{fig:session-trajectories} shows session-wise PsychEval scores underlying the main results.
The four metric families describe outcomes across a changing set of scored clients; these trajectories do not isolate learning effects from other course dynamics.

Appendix~\ref{app:case-study} traces within-state skill preferences, dialogue, and outcomes across the thirteen-session HET-860 case.

\section{Related Work}
\label{sec:related}
LLM-based counseling has progressed from multi-turn empathetic dialogue within a session \citep{liu2021esconv,chen2023soulchat} through report-grounded reconstruction \citep{zhang2024cpsycoun} and counselor-style modeling \citep{xie2025psydt} to multi-session agent systems with memory and planning \citep{wang2026muspsy,pan2026psycheval,yang2026psychagent}.
PsyEvo combines client-private skill posteriors with a shared response adapter updated at test time.

Personalization and test-time adaptation have been explored separately: LaMP and OPPU personalize generation via user histories or parameter-efficient modules \citep{salemi2024lamp,tan2024oppu}, while Self-Refine, Reflexion, and ExpeL enable iterative self-improvement through feedback loops \citep{madaan2023selfrefine,shinn2023reflexion,zhao2024expel}.
In preference optimization, DPO learns from pairwise comparisons \citep{rafailov2023dpo} and LiPO provides a general learning-to-rank framework \citep{liu2025lipo}; MICA derives mixed credit from user-state potentials \citep{zhang2026mica}.
Our LiPO objective uses SOCA support-strength weights and group normalization; the original LiPO-Lambda variant uses gain and rank-discount weights.
LLM-based evaluation can exhibit presentation bias and compressed scores \citep{zheng2023judge,liu2023geval}. Reward-model training can also produce prompt-independent reward distributions \citep{song2026rewardcollapse}, a distinct issue from the self-evaluation score compression addressed here.
SOCA constructs ordinal evidence from consistency-checked comparisons without learning a reward model.

\section{Conclusion and Limitations}
\label{sec:conclusion}
We have presented PsyEvo, a test-time learning framework that combines HBSP for client-specific skill selection, LiPO for cross-client response-policy improvement, and SOCA for candidate preferences and ordinal trajectory credit.
This separation allows individual intervention preferences to evolve alongside a shared response policy while keeping the backbone frozen.
Matched PsychEval ablations show consistent conditional component contributions across three runs; the separate SummEval study examines evidence aggregation against human judgments.

Several limitations remain.
The automatic evaluations use simulated clients and LLM judges, so their scores characterize model behavior rather than clinical efficacy.
Shared online adaptation within the scored cohort does not establish transfer to unseen clients, and individual component removals do not isolate the joint benefit of both learning paths or each SOCA mechanism.
Broader system comparisons mix evaluated and published results, while the small human-rating pilot has limited agreement and remains descriptive.
Future work will examine transfer to disjoint client cohorts, isolate the learning mechanisms with matched controls, and broaden independent human evaluation.
Extending this framework to clinician-supervised settings will also require assessment of crisis handling, privacy, and policy drift, alongside whether personalized skill selection and shared policy learning improve continuity across sessions.

\clearpage
\ifdefined\PsyEvoArxiv\else
\section*{AI Disclosure}
We used generative AI tools to polish writing, search and organize related literature, provide methodological feedback, interpret existing results, and derive and check the HBSP algebraic reformulation.
AI-assisted revisions were checked against the manuscript's equations and reported results.
The authors take responsibility for the final text, claims, and artifacts.
\fi

\bibliography{references,new-references,intro-references}
\bibliographystyle{iclr2027_conference}
\appendix
\section{Dataset Details}
\label{app:datasets-detail}
We evaluate on three datasets: PsychEval and MusPsy assess multi-session counseling quality, and SummEval measures agreement between a summary-evaluation variant of SOCA and human judgments.
Table~\ref{tab:datasets} summarizes the source corpora and the subsets used in our experiments.

\begin{table}[ht]
\centering\small
\caption{Dataset statistics and evaluation subsets.
Source size describes the released corpus; the evaluation subset is the portion used in our experiments.}
\label{tab:datasets}
\begin{tabularx}{\linewidth}{p{.17\linewidth}XXp{.17\linewidth}}
\toprule
Dataset & Source size & Evaluation subset & Reported metrics\\
\midrule
PsychEval & 369 trajectories; 2,798 sessions; 67,314 turns & 100 test clients, 20 per modality across five modalities & C-Shared, C-Spec, L-Shared, L-Spec, Overall\\
MusPsy & 1,400 training and 100 test profiles & 100 test profiles $\times$ 6 sessions (600 client--session observations) & WAI, PANAS\\
SummEval & 100 articles; 1,600 summaries from 16 systems & All 100 source documents and 1,600 summaries & Spearman $\rho$ on COH, CON, FLU, REL\\
\bottomrule
\end{tabularx}
\end{table}

\paragraph{PsychEval}
PsychEval contains 369 case-based counseling trajectories (2,798 sessions and 67,314 turns; 7.6 sessions per case and 24.1 turns per session on average) spanning cognitive behavioral, behavioral, humanistic--existential, psychodynamic, postmodernist, and integrative therapy, with a skill taxonomy of 677 meta-skills and 4,577 atomic skills.
We evaluate on a 100-client cohort across the five individual modalities, with 20 clients per modality.
The internal feedback used for learning is kept separate from external outcome evaluation (Appendix~\ref{app:judgesfull}).
Because clients can end their courses at different sessions, session-wise analyses average the scored clients at each session.

\paragraph{MusPsy}
MusPsy is a CBT-oriented synthetic corpus grounded in published case reports, with 1,400 training and 100 test client profiles averaging 6.17 sessions per client and 28.55 turns per session; each profile includes evolving client states, session goals, and counseling memories.
We evaluate all 100 test profiles over six consecutive sessions, yielding 600 client--session observations per method.
The author-released reference systems use Llama3-8B on 100 profiles; our evaluation follows the official MusPsy evaluation code and judging prompt (Appendix~\ref{app:metrics-detail}).

\paragraph{SummEval}
SummEval contains 1,600 summaries from 16 systems for 100 CNN/DailyMail articles, each rated by three experts and five crowd workers on four quality dimensions.
We evaluate all 1,600 summaries across the 100 source documents, retaining each document's 16 summaries as a group for correlation computation.

\section{Baseline Specifications}
\label{app:baselines-detail}

\subsection{Counseling Baselines}
Table~\ref{tab:counseling-main} compares PsyEvo with eleven baselines on PsychEval, including seven counseling-specific systems (CACTUS, CPsyCounX, SMILE, PsyDTLLM, PsyLLM-thinking-v3, TheraMind, PsychAgent-32B) and four general-purpose models (GPT-5.6-terra, DeepSeek-V4-Flash, Gemini-3.8-Flash, Qwen3-32B). The MusPsy figures appear separately in Table~\ref{tab:muspsy}.
PsychEval scores of CPsyCounX, PsyDTLLM, PsyLLM-thinking-v3, TheraMind, and PsychAgent-32B are cited from prior publications; all other PsychEval results are from our evaluation, and the sources of MusPsy scores are described in Appendix~\ref{app:metrics-detail}.
The comparisons include published reference scores alongside our evaluations.
Our evaluations follow the source benchmark protocols; imported scores retain their published evaluation conditions and have not been uniformly re-scored by us.
For PsychEval, we follow the public PsychAgent setup with at most 20 sessions per client and 45 counselor turns per session.
Client and counselor generation use temperatures of 0.7 and 0.9 and output limits of 512 and 8,196 tokens, respectively.
External scoring uses Gemini-3-Flash (\texttt{gemini-3-flash-preview}) and the released modality-specific evaluation methods.
The protocol covers BT, CBT, HET, PDT, and PMT, with 20 clients per modality.
The released configuration and evaluation interfaces are available at \url{https://github.com/ECNU-ICALK/PsychAgent}.
For MusPsy, the benchmark protocol uses English client profiles with dynamic states carried between six sessions; client and counselor generation both use temperature 0.7, and the original GPT-4o evaluator uses temperature 0 \citep{wang2026muspsy}.
The released profiles and simulation and evaluation prompts define the MusPsy setup independently of the PsychEval configuration.
For the author-released MusPsy references, CACTUS and SMILE denote Meta-Llama-3-8B-Instruct models fine-tuned on those datasets, rather than their original released checkpoints: the source implementation uses LoRA rank 32, $\alpha=16$, learning rate $2\times10^{-4}$, and two epochs \citep{wang2026muspsy}.
These source training settings describe the imported references, not PsyEvo's online updates.

\paragraph{Comparison provenance and interpretation.}
For PsychEval, the five cited systems listed above are published references; CACTUS, SMILE, and the four general-purpose systems are evaluated by us, alongside PsyEvo.
The component comparisons in Table~\ref{tab:components} are matched interventions within the PsyEvo scaffold; Table~\ref{tab:counseling-main} instead compares complete systems, including differences in inference scaffolding and adaptation access.
Following a common benchmark does not by itself guarantee identical model versions, judge calibration, or uncertainty for imported and newly evaluated results.
The highest-mean comparison is descriptive, not a paired significance claim; the within-scaffold interventions provide the evidence for individual component contributions.

\subsection{Summary-Evaluation Baselines}
All compared evaluators use Qwen3-32B as the backbone.
Raw Scalar selects the maximum-probability score category, while Judgment Distribution Mean (JDM) takes the expectation over score categories.
Pairwise Comparative Assessment (PCA) aggregates judgments of relative quality between two summaries \citep{liusie2024pca}.
PairS-Greedy constructs a ranking through pairwise merge-sort decisions; PairS-Beam retains multiple candidate orderings during merging to account for uncertain preferences \citep{liu2024aligning}.
BatchEval evaluates multiple candidates jointly through iterative batch-wise comparison \citep{yuan2024batcheval}.
The comparison includes all six baselines on the full SummEval dataset.
These are evaluator baselines applied to fixed SummEval outputs, rather than the 16 summarization systems that generated the dataset.
SOCA's aggregation procedure is detailed in Appendix~\ref{app:soca-aggregation}.

\section{Implementation and Reproducible Update Schedule}
\label{app:implementationfull}

The PsychEval implementation separates dialogue serving from adapter training.
It uses Qwen3-32B as the counselor, BGE-M3 for retrieval embeddings, and DeepSeek-V3.1 as the simulated client.
Four graphics processing units (GPUs) serve dialogue, and four train adapters.
The implementation maintains separate memories and skill posteriors for individual clients and accumulates cross-client preference evidence for LiPO training.
Within each run, clients across the five modalities share one response adapter and evidence buffer; their memories and skill posteriors remain client-specific.
Consequently, a session can use an adapter updated from earlier sessions of other clients in the same evaluation cohort. The reported scores assess this shared online-learning setting and do not by themselves establish transfer to a disjoint client cohort.
Each seed starts a separate run with resume disabled, and up to ten clients execute concurrently.
Table~\ref{tab:implementationfull} specifies the PsychEval full-system and component-comparison configuration.

\paragraph{Online evaluation target.}
The scored unit is a realized counseling session produced during adaptation, with an adapter version fixed when that session starts.
Accepted updates can affect subsequent sessions but cannot retroactively alter an already scored transcript.
This protocol evaluates learning within an interacting client cohort; evaluating a frozen adapter on a disjoint cohort would test a separate transfer claim.
Within a run, client outcomes can be dependent through the shared adapter even though memories and skill posteriors are private.

\begin{table}[ht]
\centering\small
\caption{Online learning configuration for the PsychEval full system and component comparisons.
Component variants retain these settings except for the specified intervention.}
\label{tab:implementationfull}
\begin{tabular}{ll}
\toprule
Setting & Value\\
\midrule
Maximum sessions / counselor turns & 20 / 45\\
Soft session target / grace & 6 / 2\\
Random seeds (PsychEval evaluation) & 7, 17, 27\\
Client / shadow temperature & 0.7 / 1.3\\
Shadow responses / total candidates & 4 / 5\\
Update eligibility interval & 5 completed sessions across clients\\
LoRA rank / batch size / gradient accumulation & 8 / 2 / 4\\
Preference pairs / minimum candidate groups & 320 / 32\\
Minimum batch share per modality & 20\% for each of the five modalities\\
Stable preference-edge strength threshold & 0.1\\
Maximum training sequence length & 1,024 tokens\\
Learning rate / preference scale & $10^{-5}$ / $\beta=0.05$\\
Maximum optimizer steps / epochs per round & 3 / 1\\
Shared skill table / backbone & Fixed / frozen\\
Preference reference & Frozen backbone, adapter disabled\\
Adapter activation & Subsequent sessions after an accepted update\\
Ordinal tiers / spread / discount & $K=3$, $s=0.5$, $\gamma=0.9$\\
HBSP prior mean / variance & $0$ / $0.25$\\
HBSP observation variance & $\sigma_b^2=1$\\
Skill-state slots & $3$ stages $\times\,3$ progress bins $\times\,6$ affect categories\\
\bottomrule
\end{tabular}
\end{table}

For counseling, the skill controller maps the dialogue to 54 slots: three counseling stages, three within-session progress bins, and six keyword-based affect categories.
The stages are problem formulation and goal setting, core intervention, and consolidation; progress bins use thresholds 0.33 and 0.66 of the configured turn budget.
The affect categories are anxiety, sadness, anger, guilt, calmness, and positive affect.
This discrete index addresses the skill table; response generation retains the dialogue context.
HBSP treats residual returns within each client-specific state--skill cell as conditionally independent observations under its Gaussian working model.
The resulting precision controls posterior sampling within this model; it is an online approximation for a policy that evolves across sessions.
The hierarchy consists of a fixed shared table and client-specific residuals; HBSP does not learn a population hyperprior online.

The realized-turn outcome $R_t\in[0,1]$ is a retrospective assessment of the client's observed emotional change, willingness to disclose, collaboration, and concrete progress after the delivered response.
The terminal reward is a weighted combination of direction-aligned, normalized PANAS, modality-specific scale, SRS, and alliance scores.
SOCA applies the same mean-preserving ordinal projection to the client's session-reward history and uses the latest projected value as $r_{\mathrm{session}}$; the first session retains its raw value.
Equation~\ref{eq:return} accumulates these credits with discounting.

Candidate generation and preference construction retain the identity of the state from which each candidate group originates.
The buffer stores the prompt, candidate identifiers, local judgments, retained preference edges, and the policy version that generated the group.
Fixing this evidence before optimization allows the loss to be reconstructed from the saved candidate group.
An external evaluation score is not substituted for a missing candidate judgment.

On PsychEval, the five-session eligibility interval counts completed sessions across clients sharing the adapter.
An optimization round uses 320 preference pairs from at least 32 candidate groups, with each of the five modalities contributing at least 20\% of the batch.
A batch that fails a requirement is returned to the buffer for further evidence accumulation.
An update proceeds through evidence collection, optimization, validation, and activation.
The active adapter remains fixed throughout each session.
At a session boundary, the implementation freezes the evidence, evaluates reference probabilities under the frozen backbone with the adapter disabled, optimizes the preference objective, and records the update outcome.
Only an accepted adapter becomes eligible for subsequent generation; a failed update leaves the previous version active.
This sequence associates each transcript with a specific policy version.

\begin{table}[ht]
\centering\small
\caption{Persistent client and counselor learning states.
Each row lists a stored object, its contents, and its role in the test-time learning cycle.}
\begin{tabularx}{\linewidth}{p{.22\linewidth}XX}
\toprule
Object & Contents & Role\\
\midrule
Memory & Recap, goals, unresolved themes & Semantic continuity across sessions\\
Skill posterior & State--skill means and precisions & Personalized selection and exploration\\
Evidence buffer & Candidates, edges, weights, versions & Preference optimization input\\
Adapter registry & Learning scope, parent version, checkpoint, activation boundary & Loading, replay, and rollback\\
\bottomrule
\end{tabularx}
\end{table}

Memory and parameter learning provide different forms of continuity.
A recap can retain a workplace trigger or a previously agreed coping exercise without changing the model weights.
Updates to the skill posterior change the relative values of interventions, while updates to the adapter change the conditional response distribution.
The separate states support memory-only, controller-only, and adapter-update evaluations.

The full system and all component variants in Table~\ref{tab:components} use the same experimental configuration described in Table~\ref{tab:implementationfull}, except for the specified component intervention.
The SOCA variant replaces consistency-checked candidate preferences with raw-score preferences and uses unprojected outcome scores for trajectory credit, while retaining grouped preference optimization.
The DPO variant retains SOCA preferences but replaces weighted group optimization with unit-weight pairwise optimization, testing the combined effect of SOCA weighting and group normalization.
The HBSP variant retains the shared skill table while removing client-private residuals and their updates.
The parameter-update variant keeps the response adapter fixed while retaining skill-posterior updates, memory, and candidate evaluation.
Other experimental settings remain unchanged across variants.
The fixed-adapter variant therefore still permits client-posterior adaptation, and the no-HBSP variant still permits adapter learning.
Neither is a control that freezes both learning paths while retaining the complete scaffold.
Likewise, the SOCA intervention jointly changes candidate preferences and realized-trajectory credit; it does not identify the separate effect of ordinal projection.

\begin{table}[htbp]
\centering\small
\caption{Scope of the matched PsychEval interventions. Memory, candidate generation, backbone, case set, and evaluation protocol are retained throughout. Each row changes the indicated component within the remaining scaffold.}
\label{tab:ablation-contract}
\begin{tabularx}{\linewidth}{p{.17\linewidth}XX}
\toprule
Variant & Changed operation & Retained learning or selection operations\\
\midrule
w/o HBSP & Remove client-private residuals and their updates & Fixed shared skill table; SOCA selection and shared-adapter learning\\
w/o LiPO & Keep the response adapter fixed & Client-posterior updates; SOCA candidate evaluation and selection\\
w/o SOCA & Use raw-score preferences and unprojected outcome credit & Client-posterior updates; grouped preference optimization\\
with DPO & Replace weighted group loss with unit-weight pair loss & SOCA preferences; client-posterior and shared-adapter updates\\
\bottomrule
\end{tabularx}
\end{table}

\subsection{Supplementary Implementation Details}
\label{app:code-details}
The supplementary method package exposes the numerical HBSP, SOCA, and population LiPO operations separately from model inference and benchmark evaluation.
It is a reference implementation of these operations, rather than an export of the experimental training system.
The following details specify its decision rules, weighting, and update lifecycle; model-specific generation, feedback prompts, and physical adapter training are supplied by a backend.

\paragraph{HBSP exploration and initialization.}
Initialization uses skill labels in the static PsychEval corpus.
The base construction maps labeled decisions to the counseling-stage, session-progress, and affect slots described above, counts skills within each state, and normalizes their frequencies.
For a state with observed labels, writing $n(s,a)$ for the skill count gives the frequency prior
\begin{equation}
Q_{\mathrm{freq}}(s,a)=\frac{n(s,a)}{\sum_{a'}n(s,a')}.
\label{eq:q-frequency}
\end{equation}
The tooling also includes a mixed-prior variant; the loaded initialization file determines which fixed table is used.
In particular, the frequency prior is a skill-selection preference score, not an empirical mean discounted return.
During evaluation, realized returns update the client-private residual against the loaded, fixed table, rather than re-estimating its corpus frequencies.
The experiments use $\tau=1.0$ and a fixed $\varepsilon_t=0.05$ throughout interaction, matching the supplied \texttt{HBSPConfig}.
The implementation's \texttt{temperature} multiplies the skill values before softmax, as in Eq.~\ref{eq:hbsp-policy}; it is not a divisor.
Posterior sampling is applied before the exploration mixture, which is uniform only over the valid skills; excluded skills receive zero probability.
The constructor copies the supplied shared table and keeps it read-only, while each new client receives separate zero-mean residuals and prior precisions.
An update changes only the selected client--state--skill cell and subtracts the corresponding shared value from the cumulative return without an additional rescaling step.
The package accepts the shared table as input; its synthetic example is not the annotated-data initialization used in the experiments.
Decision records expose the action probabilities and posterior mean, standard deviation, and draw, so posterior updates and subsequent decision-time use can be inspected separately.

\paragraph{How the skill prior combines with returns.}
\label{app:hbsp-shrinkage}
For a fixed client--state--skill cell, let $G_1,\ldots,G_n$ denote the returns assigned to its successive updates, with $m_0=0$ and fixed $Q$ and $\sigma_b^2$.
Accumulating Eq.~\ref{eq:hbsp-update} gives $\lambda_n m_n=\sigma_b^{-2}\sum_{i=1}^{n}(G_i-Q)$, and hence
\begin{equation}
Q+m_n=\frac{\lambda_0 Q+\sigma_b^{-2}\sum_{i=1}^{n}G_i}{\lambda_0+n\sigma_b^{-2}}.
\label{eq:hbsp-shrinkage}
\end{equation}
Thus the mean selection score interpolates between the loaded prior and the observed mean return; $Q$ need not itself be an empirical return estimate.
With the stated variances, its prior weight is $4/(4+n)$.
The cumulative $G_i$ and residual need not lie in the unit interval used for individual rewards.
This identity explains the update's scale behavior, not its statistical calibration: the working variance and dependence between returns still affect exploration.

\paragraph{SOCA comparison and credit interfaces.}
Qwen3-32B serves as the internal evaluator: SOCA converts its judgments into candidate preferences, which LiPO uses to update the shared response adapter.
For each candidate pair, the evaluator supplies twelve criterion-level A/B judgments, which are combined into a weighted support share.
Let $z_{\ell,ij}^{o}$ indicate whether criterion $\ell$ favors the original candidate $i$ in order $o$, and let $\omega_\ell$ denote its criterion weight.
For a complete binary judgment vector, the support share is
\begin{equation}
p_{ij}^{o}=\frac{\sum_{\ell=1}^{12}\omega_\ell z_{\ell,ij}^{o}}{\sum_{\ell=1}^{12}\omega_\ell},
\qquad o\in\{\rightarrow,\leftarrow\}.
\label{eq:criterion-support}
\end{equation}
Each presentation order is evaluated separately, with A/B choices mapped back to the original candidate identities before applying Eq.~\ref{eq:soca-graph}.
The resulting number represents weighted criterion support, not a next-token likelihood or a calibrated probability of therapeutic benefit. The edge weight in Eq.~\ref{eq:soca-graph} measures the smaller of the two directional support margins; the twelve criteria are inputs to one evaluator, not independent evaluators.
The graph constructor requires one two-order judgment for every unordered candidate pair.
Both support values refer to the same original candidate after reversing presentation order.
Agreement must be strict on the same side of $0.5$; an exact tie or order disagreement produces no edge.
To apply the configured stability floor of $0.1$, the backend must pass it explicitly, since the general graph routine defaults to a zero floor.
Equal-strength edges are ordered by a deterministic hash of the group identifier and unordered candidate indices before acyclic admission.
Pointwise scores cannot create or reverse edges; they only break equal-depth selection ties.
An empty graph therefore falls back to pointwise selection but is not admitted to preference training.
For realized outcomes, exact ties share a midrank, and a common shift followed by clipping preserves the mean; the reference solver uses bisection.
Discounted returns are accumulated backward from the supplied terminal reward.
These interfaces consume numerical feedback and do not determine which model or prompt generated it.

\paragraph{Feedback flow and optimization.}
Pointwise candidate scores and two-order comparison supports describe alternative responses at the current state, before a response is delivered.
Their role differs from realized-turn feedback, which describes the simulated client's reaction after delivery, and from terminal feedback available after the session.
The former supplies local preference edges for LiPO; the latter supplies trajectory returns for HBSP.
The preference dataset contains candidates, edge directions, and numerical weights, not gradients between groups.
LiPO differentiates policy log probabilities while holding these saved judgments fixed; SOCA does not backpropagate through evaluator calls.
The frozen, adapter-disabled preference reference specifies the denominator in Eq.~\ref{eq:lipo}; it is a distinct role from the internal evaluator that supplies comparison support.
External transcript scores are a separate reporting channel (Appendix~\ref{app:judgesfull}).

\paragraph{Population weighting and batch units.}
Each buffered item is a complete local graph identified by client and group, with modality and generating policy version retained.
Pooling these items does not introduce comparisons between different clients.
In \texttt{lipo.py}, each graph loss is divided by its retained edge count, not by the sum of support-strength weights; graph losses are then averaged equally.
Clients consequently contribute in proportion to their admitted group counts, rather than receiving equal client weights.
The listwise grouping preserves local candidate sets, while the objective itself remains a weighted pairwise logistic loss; it does not define a likelihood over complete candidate permutations.
The optional autograd implementation detaches reference log probabilities; the backend must compute them with the adapter disabled and matching response-token masks.
The reference batch sampler counts preference pairs as retained graph edges.
Under the PsychEval configuration, exactly 320 edges with at least 20\% from each of five modalities requires 64 edges per modality, together with at least 32 groups.
It selects whole graphs without truncation and leaves the buffer unchanged if no admissible batch exists.
Its deterministic selection does not reconstruct the experimental sampling order or identify the accounting unit of the separate archived training trace in Appendix~\ref{app:costfull}.

\paragraph{Session boundaries and failed updates.}
One scheduler is shared by all clients in a learning scope.
Starting a session captures the active adapter identifier; later acceptance of a new version does not change that session's identifier.
Under the PsychEval schedule, five completed sessions make an update eligible, but training still waits for an admissible batch.
Training must produce a separate candidate checkpoint, which is checked by a backend validation callback before the active reference advances.
An exception or rejection returns the reserved graphs and retains the previous active reference; returning evidence is not a rollback of arbitrary in-place model changes.
The backend must preserve the active checkpoint and implement physical loading, cleanup, and recovery.

\paragraph{Benchmark configurations.}
Benchmark-specific configurations follow the corresponding source protocols for PsychEval and MusPsy, while SummEval uses the original summary data and human annotations \citep{pan2026psycheval,yang2026psychagent,wang2026muspsy,fabbri2021summeval}.

\paragraph{MusPsy supplementary-run configuration.}
The identified MusPsy supplementary run uses CBT only, with \texttt{required\_modalities=['cbt']} and the five-modality quota disabled.
Update eligibility is checked after every two completed sessions; the configured preference-pair range is 16--160.
The LoRA rank is 8, the learning rate is $10^{-5}$, and optimization is limited to three steps per round.
These supplementary-run settings are separate from the PsychEval schedule in Table~\ref{tab:implementationfull}; their correspondence to the reported MusPsy run in Table~\ref{tab:muspsy} remains to be verified.

\section{Evaluation Metric Details}
\label{app:metrics-detail}
\subsection{PsychEval}
C-Shared covers working alliance (WAI), helping skills (HTAIS), relational authenticity (RRO), and customized clinical-quality criteria.
C-Spec uses CTRS, MITI, TES, PSC, and EFT-TFS for CBT, behavioral, humanistic, psychodynamic, and postmodernist therapy, respectively.
L-Shared covers symptoms (SCL-90), affect (PANAS), authenticity (RRO), and session alliance (SRS); L-Spec uses BDI-II, STAI, CCT, IPO, and SFBT indicators for the corresponding modalities.
Overall is the unweighted average of the four family scores.

\subsection{MusPsy}
The 12-item WAI--Short Form uses four items each for Goal (agreement on objectives), Task (agreement on counseling activities), and Bond (trust and rapport), rated on a 1--5 scale; we report the three subscale means and their unweighted average (WAI Avg).
PANAS comprises ten positive and ten negative affect items, each rated from 1 to 5; we report the mean within each group.
Higher WAI and positive PANAS and lower negative PANAS indicate better evaluator-assessed outcomes.
For MusPsy, author-released reference scores are taken from the original paper \citep{wang2026muspsy}; our evaluation averages six sessions for all 100 test clients using the official MusPsy evaluation code and judging prompt.
For the cited CACTUS and SMILE results, we retain the dataset authors' separately reported WAI average, averaged across their six published session rows.

\subsection{SummEval}
Human ratings use a 1--5 scale for coherence (COH; organization across sentences), consistency (CON; factual support from the source), fluency (FLU; sentence-level linguistic quality), and relevance (REL; coverage of salient information without redundancy).
For each dimension, Spearman's $\rho$ between automatic and human scores is computed within each source-document group and macro-averaged across groups; Mean averages the four dimension-level correlations.

\section{PsychEval Evaluation Protocol}
\label{app:archivefull}
The counselor-side metric labels in the original evaluation diagram denote working alliance (WAI), professional skill execution (HTAIS), response relevance and progression (RRO), interaction quality (Custom), and dialogue planning and continuity (DP).
These labels denote evaluation dimensions, not additional learning modules.

PsychEval evaluates counseling across repeated interactions rather than a single isolated response.
Its five therapeutic modalities are cognitive behavioral therapy (CBT), behavioral therapy (BT), humanistic--existential therapy (HET), psychodynamic therapy (PDT), and postmodernist therapy (PMT).
Counselor-shared and counselor-specific scores assess the intervention, while client-shared and client-specific scores describe the simulated client's responses.
The mean reported in the main tables assigns equal weight to the four metric-family aggregates.

Aggregation accounts for turns nested within sessions and sessions nested within clients.
For client $u$ with $K_u$ scored sessions, let $z_{uk}$ be a session-level metric.
The reported aggregate gives equal weight to clients:

\begin{equation}
\bar z_{\mathrm{client}}=\frac1U\sum_{u=1}^{U}\left(\frac1{K_u}\sum_{k=1}^{K_u}z_{uk}\right).
\end{equation}
Each client contributes one within-client mean, so longer courses do not receive greater weight.

Session-wise means are computed from the scored clients at each session; client composition can change over the course.

\subsection{Cross-Run Variability}
\label{app:seed-results}
We evaluate PsyEvo and four ablation variants on the same PsychEval case set using seeds 7, 17, and 27.
For each seed, we average session scores within each client and then equally across clients; Overall gives equal weight to the four metric families.
The reported mean gives equal weight to the three seeds, and SD is their sample standard deviation (Table~\ref{tab:seed-results}).
PsyEvo achieves the highest Overall score in each of the three runs, with a mean of 7.684 and a standard deviation of 0.028.
Across seeds 7, 17, and 27, the paired run-level gains are 0.171, 0.146, and 0.121 over DPO; 0.218, 0.169, and 0.126 over no parameter updates; 0.158, 0.153, and 0.145 over no SOCA; and 0.174, 0.136, and 0.104 over no HBSP.

\begin{table}[htbp]
\centering\small
\caption{PsychEval Overall scores across three random seeds. Mean and SD denote the across-seed mean and sample standard deviation.}
\label{tab:seed-results}
\begin{tabular}{lrrrrr}
\toprule
Variant & Seed 7 & Seed 17 & Seed 27 & Mean & SD\\
\midrule
PsyEvo w/o SOCA & 7.498 & 7.531 & 7.567 & 7.532 & 0.035\\
PsyEvo with DPO & 7.485 & 7.538 & 7.591 & 7.538 & 0.053\\
PsyEvo w/o LiPO & 7.438 & 7.515 & 7.586 & 7.513 & 0.074\\
PsyEvo w/o HBSP & 7.482 & 7.548 & 7.608 & 7.546 & 0.063\\
\textbf{PsyEvo} & 7.656 & 7.684 & 7.712 & 7.684 & 0.028\\
\bottomrule
\end{tabular}
\end{table}

For each comparison, we subtract the variant's client-level Overall score from PsyEvo's score for the same client and seed, then average these differences equally across clients and seeds.
We estimate 95\% confidence intervals using 10,000 paired client-level bootstrap resamples, retaining all three seeds and both methods for each sampled client.
These intervals summarize client-level variation conditional on the three realized online learning runs; Table~\ref{tab:seed-results} separately reports variation across runs.
Resampling client scores does not rerun the coupled adaptation process, so these intervals do not capture uncertainty over new learning schedules or new cohorts.
The consistently positive paired seed differences are useful descriptive replication evidence, but the small number of runs limits broader robustness conclusions.
Table~\ref{tab:paired-effects} reports the mean differences and intervals, which lie above zero for all four comparisons.

\begin{table}[htbp]
\centering\small
\caption{Paired differences in PsychEval Overall score. Positive differences favor PsyEvo; CI denotes the 95\% confidence interval.}
\label{tab:paired-effects}
\begin{tabular}{lrr}
\toprule
Comparison & Mean difference & 95\% CI\\
\midrule
PsyEvo $-$ with DPO & +0.146 & [0.048, 0.244]\\
PsyEvo $-$ w/o parameter updates & +0.171 & [0.085, 0.257]\\
PsyEvo $-$ w/o SOCA & +0.152 & [0.060, 0.244]\\
PsyEvo $-$ w/o HBSP & +0.138 & [0.052, 0.224]\\
\bottomrule
\end{tabular}
\end{table}

\section{Mechanism Analysis of the Three Components}
\label{app:mechanismsfull}
\subsection{Client-specific Skill Selection}
Figure~\ref{fig:hbsp-personalization} compares HBSP's top-ranked skills for 20 clients per PsychEval modality at 54 shared state-space slots.
Skill indices are defined within each modality, and comparisons are made between clients of the same modality at the same state.
Maxima tied within $10^{-12}$ are shown in gray, and a slot tied for either client is excluded from that pair's disagreement denominator.
Median disagreement is 5.6\%, 9.6\%, 13.0\%, 14.3\%, and 11.3\% for CBT, BT, HET, PDT, and PMT, respectively.
The interquartile ranges summarize the distribution over the 190 client pairs within each modality.
All 54 state-space slots are retained, and clients contribute 10--20 recorded sessions.
The analysis captures how HBSP assigns different skill rankings to individual clients under a shared state representation.

\subsection{Preference Learning and Adapter Changes}
The archived training trace records twelve accepted adapter updates over 2,080 preference pairs.
These diagnostics document parameter updates in that trace; its correspondence to the main evaluation runs and the accounting unit for pairs remain unspecified.
All twelve updates have finite gradients, and the recorded adapter norm changes from 418.83 to 421.14.
These records establish that optimization executed and changed the adapter parameters.
Counseling quality is assessed using subsequent session outcomes rather than inferred from a reduction in training loss.

Pair construction and pair optimization are separate operations.
Once a valid preferred--dispreferred pair is available, DPO can learn from the preference direction without requiring a large numerical gap between judge scores.
Score compression instead affects rules that select pairs by a fixed score threshold or multiply their losses by the score gap.
In SOCA, the weighted objective uses only the edges explicitly retained in the acyclic graph, weighted by their bidirectional-consistency strengths.

\subsection{Ordinal Signal and Delayed Credit}
\begin{figure}[ht]
\centering
\includegraphics[width=\linewidth]{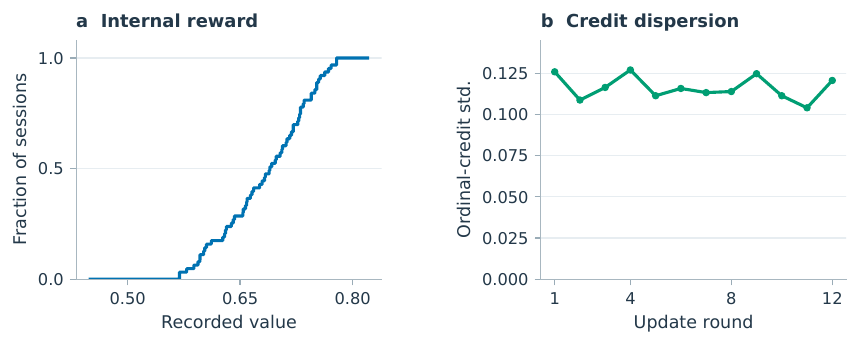}
\caption{Internal self-evaluation signal diagnostics.
(a) Empirical cumulative distribution of internal rewards over 63 sessions; (b) standard deviation of ordinal credit at each of the twelve adapter updates.
Both statistics are computed from the archived training trace.}
\label{fig:formal-signal}
\end{figure}

The trace records internal self-evaluation rewards spanning 0.569--0.779 across 63 sessions and ordinal-credit standard deviations of 0.104--0.127 across twelve updates (Figure~\ref{fig:formal-signal}), confirming that the internal learning signal is non-degenerate.
Preference agreement with human judgments is evaluated separately on SummEval (Section~\ref{sec:soca-results}).

Discounted propagation links early skill choices to later session outcomes.
Expanding the return recursion gives a weighted sum of subsequent turn credits and the terminal reward:

\begin{equation}
G_t=\sum_{j=t}^{T}\gamma^{j-t}\tilde r_j+
\gamma^{T-t+1}r_{\mathrm{session}}.
\end{equation}
An early action that builds rapport can therefore receive credit associated with later progress even when its immediate score is modest.
The discount reduces the contribution of distant outcomes.
This mechanism specifies how credit is assigned to earlier actions; isolating its effect would require a controlled intervention on the propagation rule.

\section{Reward Separation and Judge Agreement}
\label{app:judgesfull}
Reward leakage concerns the flow of evaluation information into learning.
Internal feedback is used to construct candidate preferences and trajectory returns, whereas external evaluation scores frozen transcripts.
The write-path audit checks whether external scores enter prompts, skill values, posteriors, evidence buffers, or adapter losses.
Recording call order and policy versions allows this separation to be checked.

\begin{table}[ht]
\centering\small
\caption{Feedback channels and the modules that use them.
Each row gives the unit at which a channel is recorded and the module that consumes it; only the first two channels enter learning.}
\begin{tabularx}{\linewidth}{XXX}
\toprule
Channel & Unit & Use\\
\midrule
Internal candidate judgment & Candidates in one state & LiPO preferences\\
Client-grounded reward & Realized turns and sessions & HBSP trajectory credit\\
External PsychEval judgment & Frozen transcript & Outcome reporting\\
Human ballot & Matched dialogue item & Independent quality and ranking\\
\bottomrule
\end{tabularx}
\end{table}

We assess agreement between Qwen3-32B and Gemini-3-Flash on 1,468 matched sessions from 100 clients (Table~\ref{tab:judgefull}).
We estimate 95\% confidence intervals for Spearman correlation using 2,000 client-level bootstrap resamples, retaining each sampled client's complete set of matched sessions.

\begin{table}[ht]
\centering\small
\caption{Agreement between Qwen3-32B and Gemini-3-Flash judges on 1,468 matched sessions.
Intervals are 95\% client-level bootstrap intervals (2,000 replicates) for the Spearman correlation, and offsets are mean Qwen minus Gemini scores.}
\label{tab:judgefull}
\begin{tabular}{lrrrr}
\toprule
Metric & Pearson $r$ & Spearman $\rho$ & 95\% interval & Offset\\
\midrule
C-Shared & .430 & .480 & [.412,.538] & +1.989\\
WAI & .450 & .499 & [.431,.560] & +1.398\\
C-Spec & .677 & .632 & [.531,.708] & $-.652$\\
L-Shared & .521 & .530 & [.463,.591] & +.729\\
\bottomrule
\end{tabular}
\end{table}

The judges exhibit positive rank correlations across all four reported dimensions, with Spearman correlations ranging from 0.480 to 0.632 and confidence intervals above zero.
Mean score offsets quantify their differences in calibration.
Positive judge correlations do not establish equivalent calibration, independent feedback, or clinical validity.
In particular, improvement under this simulator and these evaluators need not persist under a different simulator or a different preference source.

A separate matched reward-channel audit uses 75 sessions from 40 clients.
Client-grounded reward has Pearson correlation 0.564 and Spearman correlation 0.477 with external Gemini C-Shared.
The local supervisor checklist has lower correlations of 0.178 and 0.157, respectively.
These estimates compare different signals on a matched subset and are distinct from the 1,468-session judge-agreement estimates.

Correlation with a constant control score is undefined.
Likewise, correlation between a composite reward and one of its own components partly reflects its algebraic construction.
Neither observation establishes independence from an external judge.
The checks therefore combine non-constant matched scores with verification that external scores are not written into the training process.

\section{Human Evaluation and Longitudinal Cases}
\label{app:casesfull}
\subsection{Human Evaluation Protocol}
\label{app:human-evaluation}
To complement automatic evaluation, three experts with backgrounds in psychology rated multi-session counseling dialogues generated by PsyEvo, PsychAgent-Qwen3-32B, and CPsyCounX.
The materials cover 25 cases, with three sessions from different counseling stages per case, yielding 75 sessions.
Cases are selected at random, and the three systems are matched by case and corresponding counseling stage.
The rating platform randomly maps system identities to anonymous labels A, B, and C and randomizes display positions, while keeping the system-to-label mapping consistent across sessions of the same case.
The three raters score independently without knowing the system identities.
The platform retains the identity mapping and session metadata for matched comparisons and aggregation after scoring.
The main text reports aggregate expert ratings and the unweighted mean of six dimension-level means.
\subsection{Rating Criteria and Scale Anchors}
The evaluation interface provides dimension definitions and anchors on a 1--5 scale.
Experts rate ethics, empathy accuracy, empathy quality, interaction, and intervention for each session, and rate client improvement once after reading the three sessions of a case.
Client improvement assesses expert impressions of emotional, cognitive, or behavioral changes across those simulated sessions, rather than measured clinical change.

\begin{table}[htbp]
\centering\scriptsize
\caption{Criteria and anchors presented to expert raters for the six reported dimensions.}
\begin{tabularx}{\linewidth}{p{.12\linewidth}XXXX}\toprule
Dimension & Criterion & 1--2 & 3 & 4--5\\\midrule
Ethics & Counseling ethics and professional boundaries & Unsupported diagnoses, dangerous advice, or excessive promises & Minor boundary issues without substantive harm & Respects boundaries; clarifies diagnostic or medical limits\\
Empathy accuracy & Recognition of expressed emotion & Frequently invents unexpressed emotions & Mostly accurate, with occasional subjective inference & Accurately reflects expressed emotion\\
Empathy quality & Depth and supportive value & Little empathy or awkward, inappropriate responses & Reasonable but superficial empathy & Deep understanding and support for self-awareness\\
Interaction & Continuity, coherence, and pacing & Disconnection, repetition, or abrupt topic changes & Generally coherent, with occasional mismatch or repetition & Natural, responsive, and fluent interaction\\
Intervention & Fit to the counseling stage & Premature advice or generic guidance & Appropriate direction with imperfect timing or intensity & Gradual intervention matched to the stage and therapeutic logic\\
Client improvement & Change across three sessions & No improvement or clear deterioration & Some improvement or fluctuation & Clear, sustained positive change\\\bottomrule
\end{tabularx}
\end{table}

\subsection{Aggregation}
For the six dimensions $D_6$, let $\bar s_d$ denote the mean rating for dimension $d$.
The overall rating is
\begin{equation}
\mathrm{Overall}_6=\frac16\sum_{d\in D_6}\bar s_d.
\end{equation}
The first five means aggregate expert session ratings; client improvement aggregates expert case ratings.
The six means receive equal weight, and human-evaluation results are reported to two decimal places.

\subsection{Inter-Rater Agreement}
\label{app:human-agreement}
The agreement results reported here concern the same 25-case, three-rater evaluation as Table~\ref{tab:human-main}.
Table~\ref{tab:human-agreement} reports ordinal Krippendorff's $\alpha$ for each dimension and the two-way absolute-agreement ICC(A,3) for the three-rater mean composite, together with their reported 95\% confidence intervals.
The reported field name \emph{Evolution} is retained in the agreement table.
Dimension-level agreement is limited.
The composite reliability interval is not an interval for a paired difference between systems.
Low agreement does not alone determine whether within-rater system orderings are consistent; that question requires per-rater contrasts and paired case-level uncertainty.
The reported human comparison also does not cover the strongest general-purpose systems in the automatic evaluation.

\begin{table}[htbp]
\centering\small
\caption{Reported inter-rater agreement and reliability for the 25-case evaluation with three human raters. Ordinal $\alpha$ measures dimension-level agreement; ICC(A,3) measures absolute-agreement reliability of the three-rater mean composite.}
\label{tab:human-agreement}
\begin{tabular}{llrr}\toprule
Dimension / composite & Metric & Estimate & 95\% CI\\\midrule
Ethics & Ordinal $\alpha$ & 0.317 & [0.204, 0.455]\\
Empathy accuracy & Ordinal $\alpha$ & 0.035 & [$-0.131$, 0.171]\\
Empathy quality & Ordinal $\alpha$ & 0.153 & [$-0.040$, 0.337]\\
Interaction & Ordinal $\alpha$ & 0.106 & [$-0.057$, 0.239]\\
Intervention & Ordinal $\alpha$ & $0.350$ & [0.115, 0.585]\\
Evolution & Ordinal $\alpha$ & 0.084 & [$-0.150$, 0.235]\\
Six-dimension composite & ICC(A,3) & 0.580 & [0.398, 0.690]\\\bottomrule
\end{tabular}
\end{table}

\subsection{HET-860: Client-Specific Adaptation Across Thirteen Sessions}
\label{app:hbsp-case}
\label{app:case-study}
HET-860 is a simulated client seeking help with repeated impulsive purchases, regret, and emptiness.
We select the complete thirteen-session course to illustrate the transition from shared skill preferences to client-specific intervention selection.
In Session~1, the client describes difficulty resisting appealing purchases; by Session~13, she reports sharing a drawing with a trusted friend and finding the emptiness associated with shopping less distressing.
Drawing and self-directed reassurance increasingly become ways to respond to these urges.

\begin{figure}[htbp]
\centering
\includegraphics[width=\linewidth]{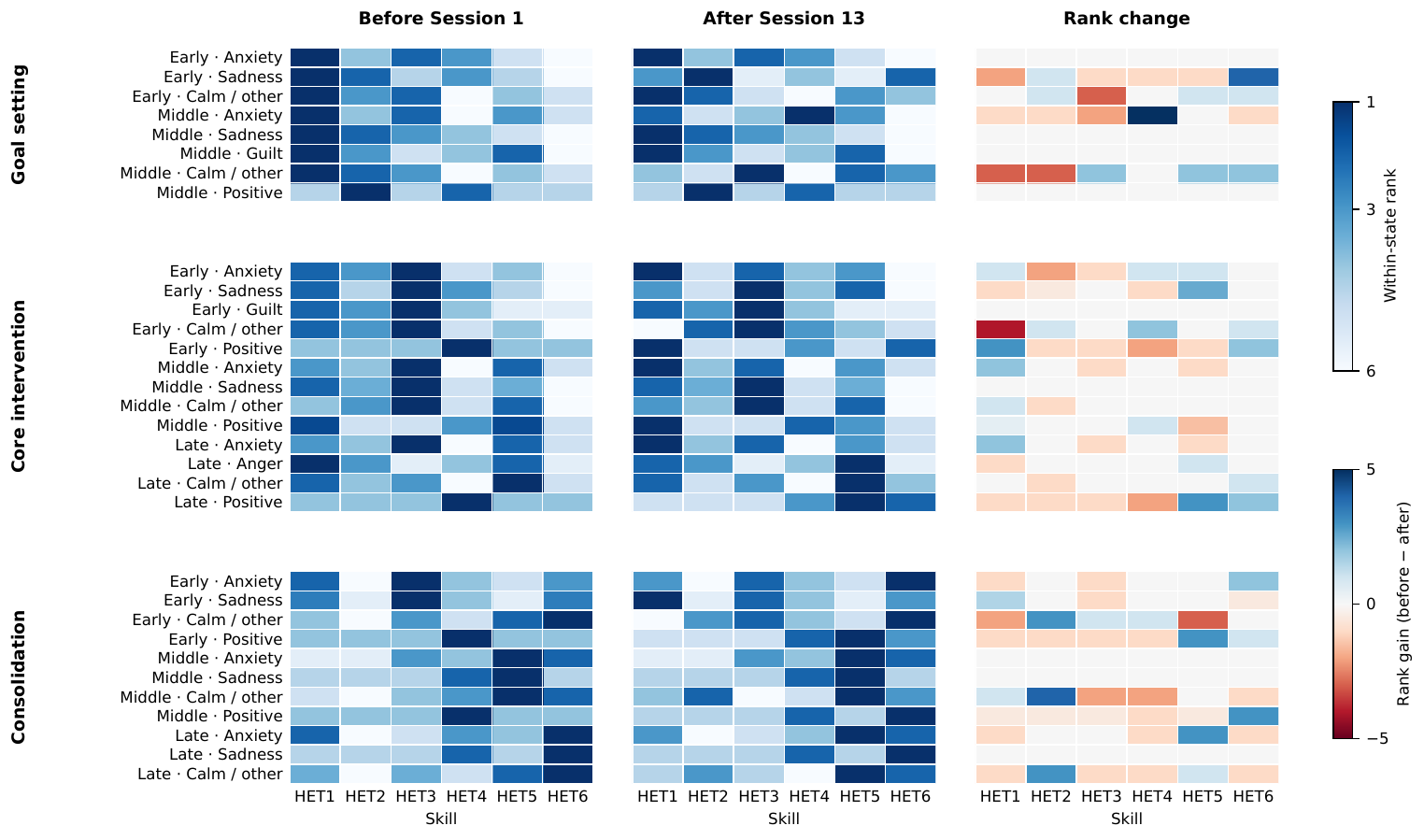}
\caption{Skill preferences before and after thirteen sessions for HET-860.
Left and middle: within-state rankings of shared values plus client-specific posterior means before Session~1 and after Session~13; darker colors indicate higher priority.
Right: initial rank minus final rank, with blue indicating increased priority, red decreased priority, and white no change; tied skills receive average ranks.
Rows cover all 32 states visited in the complete course, grouped by counseling stage. Early, Middle, and Late denote within-session progress; emotion categories are identified by keyword rules.}
\label{fig:het860-adaptation}
\end{figure}

\paragraph{Individual feedback reshapes skill preferences.}
Figure~\ref{fig:het860-adaptation} compares skill rankings before the first session and after the thirteenth.
Both snapshots evaluate all six skills at the same 32 visited states, each defined by counseling stage, within-session progress, and an emotion category.
The top-ranked skill set changes in 16 states and remains unchanged in the other 16.
Adaptation in this case is therefore state-selective: individual experience reshapes intervention priorities in some contexts while preserving shared preferences in others.

\paragraph{From shared preference to individual priority.}
For the state corresponding to core intervention, middle-session progress, and anxiety, the shared values initially favor HET3 over HET1 by 0.430.
Between the first queries of this state in Sessions~2 and~10, sixteen updates raise the HET1 client-specific posterior mean from zero to 4.404.
With shared values unchanged, the HET1-minus-HET3 value difference shifts from $-0.430$ to $+3.974$, reversing the ranking for this client.
In Session~10, selecting HET1 supplies interventions centered on acceptance, empathic listening, and nonjudgmental expression of negative feelings.
This example shows how accumulated individual feedback changes the priority of concrete interventions when the same encoded state recurs.

\paragraph{Adaptation and progress across sessions.}
The course contains 384 skill decisions and 384 posterior updates, with 282 decisions reading a nonzero client-specific residual.
The number of updated state--skill entries grows from 17 after the first session to 72 after the last.
At 66 recorded decision points, adding the posterior mean changes the highest-valued skill relative to the shared values alone.
This continual revision of the individual decision state implements non-parametric adaptation across sessions.

Over the same trajectory, external WAI increases from 6.250 to 7.500, and the mean of PANAS and SRS increases from 5.188 to 6.875.
The complete session sequence shows improvement with fluctuations: the client-shared score peaks at 7.562 in Session~9 and subsequently declines, ending above its initial value.
Table~\ref{tab:hbsp-case} presents the scores alongside posterior use and the adapter versions recorded during generation.

\begin{table}[htbp]
\centering\small
\caption{The complete thirteen-session HET-860 case. Client-shared is the mean of external PANAS and SRS scores. Nonzero reads count decisions that access a nonzero posterior mean; cells count cumulatively updated state--skill entries. Adapter versions are those recorded during generation.}
\label{tab:hbsp-case}
\begin{tabular}{rrrrrr}\toprule
Session & WAI & Client-shared & Nonzero reads & Cells & Adapter\\\midrule
1 & 6.250 & 5.188 & 0/28 & 17 & $v_{0}$\\
2 & 6.042 & 4.875 & 0/25 & 31 & $v_{1}$\\
3 & 7.292 & 4.938 & 21/29 & 37 & $v_{1}$\\
4 & 7.917 & 6.938 & 30/33 & 40 & $v_{2}$\\
5 & 7.917 & 6.188 & 29/29 & 41 & $v_{3}$\\
6 & 7.083 & 5.312 & 34/35 & 44 & $v_{4}$\\
7 & 7.708 & 6.500 & 27/27 & 45 & $v_{5}$\\
8 & 6.667 & 6.062 & 23/23 & 45 & $v_{5}$\\
9 & 6.667 & 7.562 & 34/34 & 46 & $v_{6}$\\
10 & 7.917 & 6.500 & 33/35 & 48 & $v_{6}$\\
11 & 6.667 & 6.875 & 0/23 & 61 & $v_{7}$\\
12 & 6.875 & 6.438 & 21/25 & 65 & $v_{11}$\\
13 & 7.500 & 6.875 & 30/38 & 72 & $v_{11}$\\
\bottomrule
\end{tabular}
\end{table}

\subsection{A Client-specific Course: CBT-412}
The CBT-412 archive follows a simulated retail worker whose anxiety and self-devaluation are triggered when colleagues receive praise.
The client repeatedly thinks that they are inferior to others, with accompanying chest tightness and avoidance of a supervisor.
This case illustrates how changes in the client state connect memory, skill selection, and response-policy learning across five saved sessions.

The first session identifies a trigger--thought--emotion--avoidance chain.
When imagining the supervisor approaching, the client reports anxiety of at least eight on a ten-point scale and agrees to begin with a familiar calming image.
Later summaries retain the workplace trigger, bodily cues, and the belief that external recognition determines personal value.
These retained details allow subsequent interventions to address the same difficulty without repeating a generic intake.

The recorded skill sequence progresses from emotion regulation toward trigger monitoring, cognitive distancing, and self-supportive language.
By the fifth session, the summary records that the client recognizes the recurring thought and remains in the situation rather than immediately withdrawing.
Residual discomfort remains in the profile.
This progression is documented in the narrative, with memory and both learning paths present throughout the course.

\begin{table}[ht]
\centering\small
\caption{Recorded state changes for client CBT-412.
The first three rows track session-wise state; the final two report the archived ordinal tiers and expanded credit.}
\begin{tabularx}{\linewidth}{p{.25\linewidth}X}
\toprule
Record & Observed sequence\\
\midrule
Sessions & 1, 2, 3, 4, 5\\
Updated posterior cells & 14, 30, 34, 42, 43\\
Adapter versions & $v_0,v_0,v_1,v_2,v_3$\\
Archived ordinal tiers & $[0,0,1,2,1]$\\
Archived expanded credit & $[0.241,0.241,0.741,1.000,0.741]$\\
\bottomrule
\end{tabularx}
\end{table}

Across the five sessions, the number of updated posterior cells grows from 14 to 43, while the active adapter advances from $v_0$ to $v_3$.
Together with the retained workplace context, these records illustrate how client-specific skill adaptation and response-policy updates accompany a continuous counseling course.

\subsection{Failure Cases and Memory Dependence}
A fully tied score group has no recoverable strict order.
One archived run records 725 rewards equal to 0.5 across 51 cases.
These observations support recording variance and skip decisions instead of assigning arbitrary preferences.
They also distinguish a failure of the score format from successful ordinal reconstruction when the inputs remain distinguishable.

Cross-session memory can preserve both useful context and earlier errors.
If a recap incorrectly records a goal or belief, subsequent skill selection and candidate evaluation may reinforce that mistake.
A correction from the client should therefore be incorporated into the retained profile with its source preserved.
Memory ablations should compare absent memory, fixed summaries, and revisable summaries before session trends are attributed to HBSP or LiPO.

\section{Compute, Deployment, and Safety}
\label{app:costfull}
The principal online overhead consists of shadow generation, local evaluation, and adapter optimization.
The instrumented training trace records 2,080 pairs and 2,217.6 seconds of trainer time over twelve rounds, or 36.96 minutes in total.
The mean trainer time in this trace is 3.08 minutes per round on four trainer GPUs.
This trainer-only measurement excludes dialogue serving and shadow generation and does not establish response latency or end-to-end session time.

Let $N_s$ denote the number of additional shadow responses generated at each of $T$ opportunities.
Let $C_g$ be the generation cost per shadow, $C_j$ the amortized local evaluation cost per candidate, including pointwise scoring and pairwise comparisons, and $C_o(P)$ the optimization cost for $P$ pairs.
If the delivered response is evaluated together with the shadows, the incremental cost over ordinary single-response generation is approximately

\begin{equation}
C_{\mathrm{extra}}\approx T\left[N_sC_g+(N_s+1)C_j\right]+C_o(P).
\end{equation}
Distinguishing shadow responses from the full candidate set prevents double-counting the cost of generating the delivered response.
The instrumented configuration uses $N_s=4$ shadows plus the delivered response, giving $N=N_s+1=5$ candidates.
A saved rank-8 adapter occupies approximately 79.8 MB, allowing the large backbone to remain shared.
Storage includes client-specific memories and posteriors, cross-client preference evidence, and optimizer states and adapter versions maintained for each learning scope.

Synchronous activation allows update timing to be audited but can delay the next session.
An asynchronous service could keep the previous verified adapter active while preparing a replacement, whereas an uncertainty-triggered service could reduce generation and training frequency.
These are deployment alternatives to the measured synchronous configuration.
Their effects on latency and the continued use of an earlier policy version require separate measurement.

\begin{table}[ht]
\centering\small
\caption{Deployment risks and candidate controls.
Each row names a risk, the path through which it could arise under test-time learning, and a control whose effectiveness remains to be evaluated.}
\begin{tabularx}{\linewidth}{p{.23\linewidth}XX}
\toprule
Risk & Failure path & Control\\
\midrule
Crisis handling & Engagement reward competes with escalation & Independent crisis routing and professional oversight\\
Policy drift & Repeated local updates shift behavior & Fixed checks, versioning, rollback\\
Preference poisoning & Adversarial clients inject misleading preference evidence & Per-client isolation, client-balanced loss, edge caps, validation before activation\\
Memory error & Incorrect summaries persist & Source-linked correction\\
Privacy & Sensitive information in memory or adapters & Provenance, access control, deletion and unlearning evaluation\\
State/version routing & Wrong client memory or counselor version loaded & Identity checks and activation records\\
\bottomrule
\end{tabularx}
\end{table}

Automatic evaluations use simulated clients, and human ratings assess interaction quality. Clinical efficacy and safety require prospective evaluation with professional oversight, privacy protections, and crisis-handling assessment.

\section{Update Schedule and Ordinal Properties}
\label{app:ordinal-properties}
Algorithm~\ref{alg:lifecycle} gives the complete online update schedule.
This appendix describes the conditional properties of the ordinal representation and its learning signal; implementation details and version records are provided in Appendix~\ref{app:implementationfull}.

\begin{proposition}
For a strictly increasing function $f$, $R_i>R_j$ if and only if $f(R_i)>f(R_j)$.
Consequently, the tie-aware ranks and tiers used for realized-trajectory projection are invariant to a strictly increasing transformation of the turn scores.

\end{proposition}
The proof follows directly from strict monotonicity.
Clipping after ordinal expansion preserves the mean because the implementation solves for a common shift, although the projected distances need not remain invariant.
Quantization can also introduce ties, and a non-monotone evaluator can reverse an ordering.
The proposition concerns the realized-trajectory representation; SOCA candidate edges are instead admitted by the two-order consistency rule in Eq.~\ref{eq:soca-graph}.

For an accepted edge, the derivative with respect to the utility difference $d_{ij}=h_i-h_j$ is $-w_{ij}\sigma(-d_{ij})$.
It is nonzero for positive weight and finite utility difference.
This property applies in score space; the parameter gradient also depends on the model Jacobian.
Standard DPO likewise learns from preference labels and does not require a numerical reward gap once a valid pair exists.

\section{Evidence Aggregation and Additional Results}
\subsection{Supplementary MusPsy Comparison}
\label{app:muspsy-results}
Table~\ref{tab:muspsy} reports the supplementary MusPsy figures for six sessions over 100 profiles.
The main PsyEvo run has not yet been linked to the supplementary configuration in Appendix~\ref{app:code-details}; these figures therefore do not serve as evidence for the core method comparison.
The CACTUS and SMILE rows are published references, using the dataset authors' separately reported WAI averages; the other baseline rows do not yet have a complete row-level source mapping in the available record.
The numerical comparison is retained for completeness, without a superiority or robustness claim.

\begin{table}[htbp]
\centering\small
\caption{Supplementary MusPsy figures. WAI and positive PANAS favor higher values; negative PANAS favors lower values. The main PsyEvo configuration linkage remains unresolved, so these point estimates are not used to establish the paper's principal experimental claims.}
\label{tab:muspsy}
\setlength{\tabcolsep}{3pt}
\begin{tabular}{@{}lrrrrrr@{}}
\toprule
Method & WAI Goal & WAI Task & WAI Bond & WAI Avg & PANAS Pos. & PANAS Neg.\\
\midrule
CACTUS & 4.125 & 3.713 & 4.427 & 4.085 & 3.035 & 2.333\\
CPsyCounX & 4.352 & 4.239 & 4.639 & 4.410 & 3.325 & 1.317\\
SMILE & 3.582 & 3.647 & 4.257 & 3.827 & 2.695 & 2.875\\
PsyDTLLM & 4.327 & 4.219 & 4.595 & 4.380 & 3.118 & 1.231\\
PsyLLM-thinking-v3 & 4.406 & 4.314 & 4.688 & 4.469 & 3.547 & 1.248\\
TheraMind & 4.409 & 4.312 & 4.667 & 4.463 & 3.438 & 1.236\\
PsychAgent-32B & 4.467 & 4.375 & 4.695 & 4.512 & 3.408 & 1.134\\
GPT-5.6-terra & 4.522 & 4.463 & 4.766 & 4.584 & 3.531 & 1.210\\
DeepSeek-V4-Flash & 4.499 & 4.418 & 4.739 & 4.552 & 3.479 & 1.188\\
Gemini-3.8-Flash & 4.511 & 4.467 & 4.768 & 4.582 & 3.305 & 1.257\\
Qwen3-32B & 4.486 & 4.401 & 4.741 & 4.543 & 3.462 & 1.136\\
\textbf{PsyEvo} & \textbf{4.567} & \textbf{4.551} & \textbf{4.799} & \textbf{4.639} & \textbf{3.997} & \textbf{1.057}\\
\bottomrule
\end{tabular}
\end{table}

\subsection{Multi-evidence Aggregation Beyond Counseling}
\label{app:soca-aggregation}
We use SummEval to evaluate a summary-ranking variant of SOCA against human judgments \citep{fabbri2021summeval}.
For summaries of the same source document, SOCA combines pointwise scores, relative preferences from pairwise comparisons, and accepted repair evidence.
Pointwise scores assess individual candidate quality, pairwise comparisons assess relative quality, and repairs provide evidence of correctable defects.

For candidate $i$ and quality dimension $d$, let $q_{i,d}$ be the expected pointwise score, $v_{i,d}$ the mean pairwise win rate across the two presentation orders, and $r_{i,d}$ the accepted-repair signal.
Within each source-document group, the fused score is
\begin{equation}
f_{i,d}=z(q_{i,d})+\alpha_d z(v_{i,d})-\beta_d z(r_{i,d}),
\qquad z(x_i)=\frac{x_i-\bar x}{\mathrm{sd}(x)+10^{-8}}.
\label{eq:summeval-fusion}
\end{equation}
The pointwise channel has unit weight; $\alpha_d$ and $\beta_d$ control the contributions of pairwise and repair evidence for dimension $d$.
We set $\alpha_d=\beta_d=1$ for every dimension a priori and keep these coefficients fixed throughout evaluation.
Equal coefficients apply after within-document channel standardization; they are not selected by maximizing the reported human correlations.
The pointwise expectation is $q_{i,d}=\sum_{k=1}^{5}k\,p_{i,d}(k)$.
The repair signal is the mean probability that the accepted revision is preferred in the two presentation orders, and is zero when no repair is accepted.
Its negative coefficient treats an accepted improvement as evidence of a defect in the original summary.
Candidates are ranked by decreasing $f_{i,d}$; a constant channel contributes zero after standardization.
SummEval measures agreement between the fused scores and human rankings, rather than the counseling graph selector or trajectory-credit mechanism.
The mean correlation differences are descriptive: no paired document-level uncertainty is reported, so the observed margin is not a significance claim.
In counseling, the SOCA ablation tests the combined contribution of consistency-checked candidate preferences and ordinal trajectory credit (Section~\ref{sec:soca}).

The repair module proposes a local edit and checks its validity.
The original and repaired summaries are compared in both presentation orders to check preference consistency.
Only accepted repairs contribute to the repair channel; pointwise and pairwise evidence remain available for every candidate.

JDM uses the expectation of the score distribution, whereas Raw Scalar uses its maximum-probability category.
PCA evaluates relative candidate quality through pairwise comparisons \citep{liusie2024pca}.
The evidence-channel ablations are reported in Table~\ref{tab:soca-ablation}.

\end{document}